\documentclass{article} 
\usepackage{iclr2026_conference,times}

\iclrfinalcopy

\usepackage{amsmath,amsfonts,bm}

\def\eqref#1{equation~\ref{#1}}

\def\1{\bm{1}}

\DeclareMathAlphabet{\mathsfit}{\encodingdefault}{\sfdefault}{m}{sl}
\SetMathAlphabet{\mathsfit}{bold}{\encodingdefault}{\sfdefault}{bx}{n}

\usepackage{hyperref}
\usepackage{url}
\usepackage{xcolor}
\usepackage{svg}
\usepackage{graphicx}
\usepackage{booktabs}
\usepackage{multirow}
\usepackage{makecell} 
\usepackage{amssymb} 

\definecolor{linkblue}{HTML}{1F5AA6}
\definecolor{citeblue}{HTML}{1F5AA6}
\definecolor{urlblue}{HTML}{0000EE}

\hypersetup{
    colorlinks=true,
    linkcolor=linkblue,   
    citecolor=citeblue,   
    urlcolor=urlblue,     
    pdfauthor={Yihao Wang},
    pdftitle={ROSE: Benchmarking the Perception-to-Action Gap in Multimodal Models}
}

\definecolor{resourceblue}{HTML}{0000EE}

\title{
ROSE: Benchmarking the Perception-to-Action Gap in Multimodal Models
}

\newcommand{\emailmark}[1]{%
  \href{mailto:#1}{\textsuperscript{\texttt{@}}}%
}

\author{
\textbf{Yihao Wang\textsuperscript{1}\emailmark{wangyh357@mail2.sysu.edu.cn}},
\textbf{Zijian He\textsuperscript{1}\emailmark{hezj39@mail2.sysu.edu.cn}},
\textbf{Jie Ren\textsuperscript{2}\emailmark{renjie@snnu.edu.cn}},
\textbf{Keze Wang\textsuperscript{1,\textdagger}}
\\[0.5em]
\textsuperscript{1}Sun Yat-sen University
\quad
\textsuperscript{2}Shaanxi Normal University
\\[0.4em]
{\small
\textsuperscript{\textdagger}Corresponding author:
\href{mailto:kezewang@gmail.com}{\texttt{kezewang@gmail.com}}
}
}

\begin{document}

\maketitle

\pagestyle{fancy}
\fancyhead{}
\fancyhead[L]{\small Preprint}
\fancyhead[R]{\small June 2026}
\renewcommand{\headrulewidth}{0.4pt}
\thispagestyle{fancy}

\vspace{-2.5em}
\begin{center}
\small

\href{https://xbdxwyh.github.io/ROSE-v0.1/}{%
  \raisebox{-0.18\height}{%
    \includegraphics[height=1.15em]{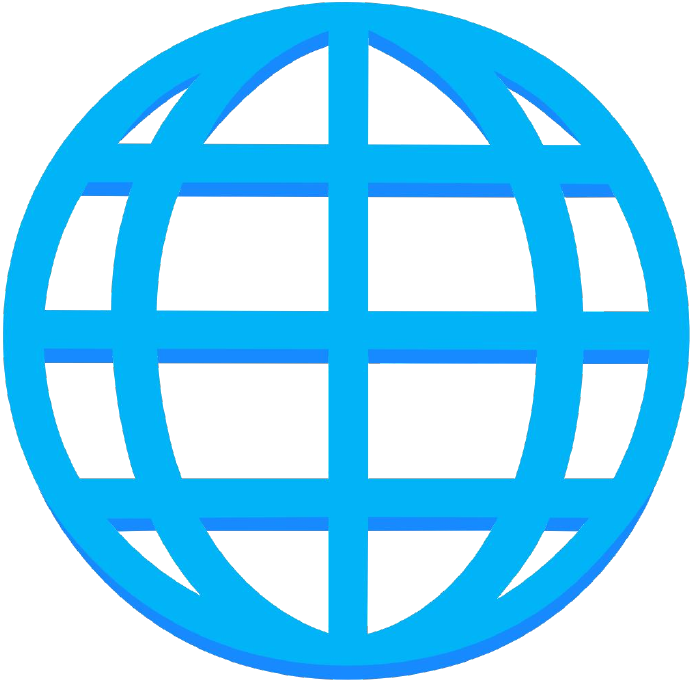}%
  }%
  \hspace{0.30em}%
  \textcolor{resourceblue}{Project Page}%
}
\hspace{1.8em}
\href{https://github.com/xbdxwyh/ROSE-v0.1}{%
  \raisebox{-0.18\height}{%
    \includegraphics[height=1.15em]{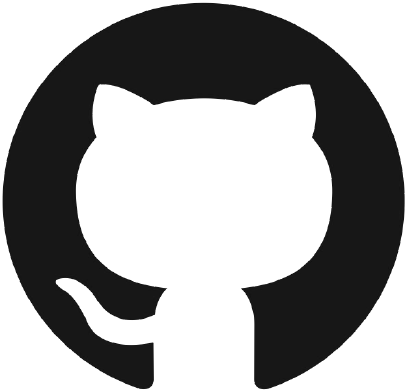}%
  }%
  \hspace{0.30em}%
  \textcolor{resourceblue}{Evaluation Code}%
}
\hspace{1.8em}
\href{https://huggingface.co/datasets/sysuwyh357/ROSE-v0.1}{%
  \raisebox{-0.18\height}{%
    \includegraphics[height=1.20em]{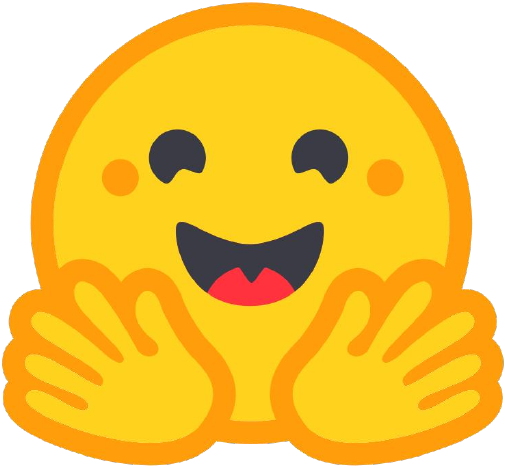}%
  }%
  \hspace{0.30em}%
  \textcolor{resourceblue}{Dataset}%
}

\end{center}


\begin{abstract}
Multimodal large language models (MLLMs) are increasingly expected to act on
visual information, yet the same scene may require different actions under
different task contexts.
How reliably can a model turn the same visual evidence into the action required
by the current context? 
To answer this question, we introduce \textsc{ROSE}
(\textbf{R}eference-conditioned \textbf{O}ddity and
\textbf{S}ymbolic \textbf{E}xecution), a controlled benchmark that holds the
visual scene fixed while varying region constraints and required symbolic
outputs.
Through coupled counting and coordinate-action tasks, \textsc{ROSE} tests
whether models can infer an implicit majority reference and act on the
resulting fine-grained visual evidence under changing contexts.
Across nine recent MLLMs, performance drops by as much as 44.5 percentage
points from counting-oriented tasks to region-conditioned action, despite
98.8\% human performance.
The gap persists on paired scenes and regions for which the same model returns
the correct count, while global-click and matched local controls show that
coordinate grounding explains only part of the loss, revealing a distinct,
model-dependent bottleneck in turning shared visual evidence into
context-specific actions.
\end{abstract}

\section{Introduction}

Recent multimodal large language models (MLLMs), such as GPT
models~\citep{achiam2023gpt,hurst2024gpt},
Gemini~\citep{team2023gemini}, and Qwen
models~\citep{bai2025qwen3,team2026qwen3}, have shown remarkable progress in
visual perception and reasoning~\citep{yi2026multimodal}.
They can now describe images and answer visual
questions~\citep{Yang_2025_CVPR}, localize
objects~\citep{bai2025qwen3,dong2026refadv,Xu_2026_CVPR}, interpret
charts~\citep{lu2026domaincqa,kondic2026chartnet} and
documents~\citep{yu2025bbox,yu2026minicpm}, and solve increasingly complex
multimodal tasks~\citep{chen2026cogflow,xie2026m,huti2026visual}.
As these models are increasingly expected to interact with visual
environments~\citep{pmlr-v270-kim25c,dang2025rynnec,dang2026rynnbrain}, an
important next step is to move beyond recognizing what is present toward
deciding how to act under task-specific visual contexts.
Yet this transition is difficult to assess in a principled way: a correct
action is not a direct by-product of recognition, but a context-dependent
commitment to what matters and what should be done.

Recent benchmarks have pushed MLLM evaluation in several complementary
directions.
Unified suites such as MME-Unify (MME-U) provide standardized evaluation
across multimodal understanding and generation~\citep{xie2025mmeunify}.
Meanwhile, OmniSpatial targets comprehensive spatial reasoning, VisuLogic
emphasizes vision-centric logical reasoning, and VGRP-Bench studies structured
visual grid puzzles~\citep{jia2025omnispatial,xu2025visulogic,ren2025vgrp}.
Embodied benchmarks and vision-language-action systems further move evaluation
toward navigation, manipulation, long-horizon planning, and executable
control~\citep{yang2025embodiedbench,pmlr-v270-kim25c}.
However, these settings are not designed to specifically isolate the
perception-to-action interface: unified suites aggregate heterogeneous tasks,
visual reasoning benchmarks typically evaluate a fixed question or puzzle
specification for each image, and embodied settings couple visual decisions
with planning, control, and environment interaction.
This raises a more focused question:
\textit{can an MLLM convert a visual interpretation into the exact action
required by the current task context while the underlying scene remains
unchanged?}

\begin{figure*}[!t]
    \centering
    \includegraphics[width=\textwidth]{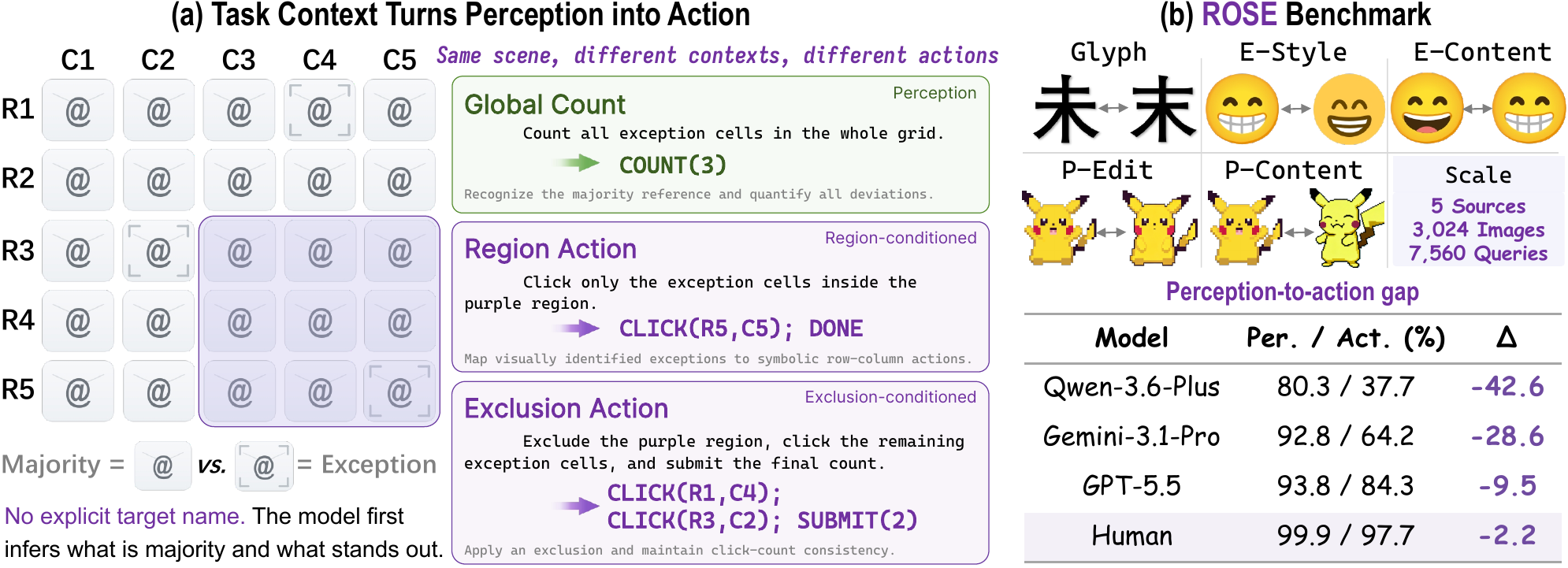}
    \caption{
    Overview of the ROSE benchmark.
    (a) Given a grid scene without an explicit target name, the model first
    infers the majority reference and reasons about the exception cells, then
    produces different formal outputs under different task contexts.
    (b) ROSE consists of five fine-grained visual sources and reveals a
    perception-to-action gap in MLLMs, where
    $\Delta=\mathrm{Act.}-\mathrm{Per}$.
    }
    \label{fig:rose_overview}
\end{figure*}

To study this question, we introduce \textbf{ROSE}
(\textbf{R}eference-conditioned \textbf{O}ddity and
\textbf{S}ymbolic \textbf{E}xecution), a controlled benchmark for
context-conditioned visual action in MLLMs.
As illustrated in Figure~\ref{fig:rose_overview}, ROSE holds the visual scene
fixed while varying the relevant region and required output operation.
Each scene presents a grid of visually similar elements without an explicit
target name, requiring the model to infer a scene-internal majority reference
and reason about sparse exceptions.
The same visual evidence is then queried through global, region-conditioned,
and exclusion-based tasks, with outputs ranging from compact counts to exact
coordinate actions.
This scene-level coupling creates controlled within-scene comparisons in which
the model must change what is selected and how it is expressed, rather than
solve an unrelated task on a different image.

ROSE is designed to be both controlled and diagnostic.
Its five visual sources vary the carrier of the fine-grained distinction while
preserving the same scene-to-task protocol, reducing reliance on a particular
semantic category or visual cue.
Counting is treated as a lower-demand behavioral readout rather than proof of
exact localization, while paired action tasks test whether the same evidence
supports region-sensitive selection and coordinate-level execution.
The strict formal-output protocol additionally separates grammar compliance
from exact task success.
Together, these controls make it possible to distinguish failures associated
with cardinality readout, coordinate localization, context-conditioned
selection, and formal action execution.

We evaluate nine recent MLLMs on ROSE and include a single trained human
annotator as a solvability reference.
The human reference achieves 98.8\% average PASS, while GPT-5.5 reaches 92.2\%,
Gemini-3.1-Pro reaches 79.4\%, and the remaining models range from 14.3\% to
50.3\%.
More importantly, current models exhibit a strongly model-dependent
counting-to-action gap, with performance dropping by as much as 44.5 percentage
points when compact counting readouts are replaced by region-conditioned
actions.
This gap persists on paired scenes and regions where the same model answers the
corresponding counting query correctly.
Global-click and exactly matched local controls further show that coordinate
grounding explains only part of the loss: some models can recover the correct
cardinality, and sometimes the global coordinates, but still fail to construct
the exact target set required by the current context.
High grammar-valid rates alongside much lower PASS scores further confirm that
these failures cannot be attributed to output formatting alone.

Our contributions are threefold:
\begin{itemize}
    \item We formulate reference-conditioned visual action, a controlled
    setting that holds visual evidence fixed while varying the relevant context
    and required symbolic readout.

    \item We build ROSE, comprising 1,512 scenes, 3,024 images, and 7,560 task
    instances across five fine-grained visual sources and five coupled task
    templates.

    \item We provide a diagnostic evaluation of nine recent MLLMs, including
    global-click and exactly matched local controls that separate output
    validity, coordinate localization, and context-conditioned action.
\end{itemize}

\section{Related Work}
\label{sec:related_work}

\paragraph{Multimodal visual reasoning benchmarks.}
The rapid development of MLLMs has motivated broad evaluation suites covering visual perception, knowledge, reasoning, and generation.
Benchmarks such as MME, MM-Vet, MMMU, and MME-Unify assess increasingly diverse and integrated multimodal capabilities~\citep{fu2023mme,yu2024mmvet,yue2024mmmu,xie2025mmeunify}.
More targeted benchmarks seek to reduce linguistic shortcuts and place greater emphasis on vision-centric reasoning.
BLINK probes fundamental visual perception skills, VisuLogic evaluates visual logic across several reasoning categories, OmniSpatial focuses on higher-order spatial cognition, and VGRP-Bench studies rule-based reasoning over structured grid puzzles~\citep{fu2024blink,xu2025visulogic,jia2025omnispatial,ren2025vgrp}.
These benchmarks reveal substantial weaknesses in visual perception and structured reasoning, but they generally evaluate each image under a fixed question or puzzle specification.
ROSE instead derives multiple coupled tasks from the same visual scene and varies both the relevant region and the required output operation.
This design enables a controlled, within-scene evaluation of whether a visual interpretation transfers across task contexts.

\paragraph{Fine-grained discrepancy and context-dependent perception.}
Several recent benchmarks more directly examine whether multimodal models can detect subtle visual differences.
SalBench evaluates low-level visual saliency through odd-one-out detection and referring variants, while OddGridBench uses controlled grid images containing a single anomalous icon that differs in attributes such as color, size, rotation, or position~\citep{dahou2025salbench,weng2026oddgridbench}.
These works provide important evidence that seemingly simple visual discrepancies remain challenging for current models.
ROSE is complementary but targets a different capability.
Its exceptions are defined relative to an implicit majority reference, may occur at multiple locations, and are drawn from glyph-, emoji-, and pixel-art-level variations rather than only parameterized low-level attributes.
More importantly, detecting the global exception set is only the perceptual basis of the task: the model must subsequently filter that set under numeric, visual, or exclusion-based regions and convert the result into an exact symbolic answer.

Context-sensitive visual understanding has also been explored by ConTextual and CODIS, which require models to use textual context to interpret text-rich or inherently ambiguous images~\citep{wadhawan2024contextual,luo2024codis}.
In these benchmarks, context primarily changes or disambiguates the semantic interpretation of an image.
In ROSE, by contrast, the underlying visual evidence and majority relation remain fixed; context determines which members of an already established exception set are relevant and what formal action should be executed.
ROSE therefore focuses on context-conditioned visual selection rather than contextual semantic disambiguation.

\paragraph{Visual grounding and action.}
Visual grounding connects language descriptions to spatial regions and provides an important foundation for visually guided action.
Classical referring-expression tasks and recent challenging variants such as Ref-Adv evaluate whether models can localize a target specified through language~\citep{dong2026refadv}.
GUI grounding further maps textual instructions to precise screen coordinates, while embodied and vision-language-action benchmarks evaluate navigation, manipulation, and longer-horizon interaction in visual environments~\citep{gou2024uground,yang2025embodiedbench,pmlr-v270-kim25c}.
These settings bring multimodal models closer to executable action, but the target is typically named or described in the instruction, and performance may depend jointly on semantic grounding, planning, control, memory, and environment feedback.
ROSE removes these confounding factors: the target identity must be inferred from the current visual scene, the action space is an automatically verifiable set of grid coordinates, and scene-coupled counting and clicking tasks isolate the transition from visual reference inference to context-conditioned symbolic execution.

\section{ROSE Benchmark Design}
\label{sec:benchmark}

\subsection{Benchmark Principle and Formalization}
\label{sec:benchmark_overview}
\label{sec:task_definition}

ROSE is designed to turn the perception-to-action gap into a controlled
\emph{readout problem}.
Rather than comparing tasks built from different images, it holds the visual
evidence fixed and changes only what must be extracted from that evidence and
how it must be expressed.
Multiple counting-oriented and action-oriented tasks are derived from the same
scene, with different region contexts and symbolic output requirements.
This makes it possible to ask whether a visual interpretation that supports a
simple cardinality judgment also supports exact, context-dependent action,
without confounding the comparison with a change in the underlying image.

The grid is therefore not merely a convenient presentation format.
Its repeated structure creates a scene-internal reference without explicitly
naming the target: the dominant element defines what is visually normal, and
the sparse deviations are meaningful only relative to that reference.
At the same time, the discrete cells provide a common coordinate system over
which the same inferred exception set can be read out in different ways---as a
global count, a region-restricted count, an exact coordinate set, or an
exclusion-conditioned action.
Within each scene, the majority element, exception identity, and global
exception locations remain fixed; the benchmark intervenes only on the
relevant context and the required output operation.
In this sense, ROSE converts one visual scene into a family of controlled
behavioral probes of how perception is transformed into action.

Formally, let $\mathcal{G}$ denote the set of grid cells and let $v_{r,c}$ be
the visual element rendered at cell $(r,c)$.
The dominant visual pattern defines an implicit majority reference $v^\star$,
and the global exception set is

\begin{equation}
\mathcal{O}
=
\{(r,c)\in\mathcal{G}:v_{r,c}\neq v^\star\}.
\label{eq:global_exception_set}
\end{equation}

Because $v^\star$ is inferred from the image rather than provided in text,
$\mathcal{O}$ is defined relative to the visual structure of the current scene.
A task further specifies a permitted region
$\mathcal{R}\subseteq\mathcal{G}$, yielding the context-specific target set

\begin{equation}
\mathcal{T}
=
\mathcal{O}\cap\mathcal{R}.
\label{eq:context_target_set}
\end{equation}

ROSE then varies how this same target set must be read out.
Depending on the task, the model must return its cardinality
$|\mathcal{T}|$, its exact coordinate set $\mathcal{T}$, or both the
coordinate set and a consistent submitted count.
The visual evidence and target definition are therefore shared, while the
required output ranges from a compact count to explicit coordinate-level
action.

A correct count is treated as lower-demand behavioral evidence, not as proof
that every target has been precisely localized.
By comparing independently queried tasks derived from the same scene, ROSE
measures how consistently a model can convert shared visual evidence into
different context-dependent outputs.
Additional matched controls that further separate counting, localization, and
context-conditioned selection are described in
Appendix~\ref{app:controlled_bridges}.

\subsection{Coupled Task Suite}
\label{sec:scene_task_instantiation}
\label{sec:region_conditions}

For each curated visual pair, we construct a scene by assigning one element as
the majority and the other as the exception.
The scene fixes the grid layout, rendered assets, and global exception
locations.
ROSE then derives a sequence of coupled tasks that progressively changes what
must be extracted from this shared visual evidence and how explicitly it must
be acted upon.

\paragraph{From global readout to context-conditioned selection.}
We begin with \textbf{T1: Global counting (G-Cnt)}, which asks for the number
of exception cells in the full grid.
Counting provides a compact and exactly verifiable readout of the scene while
avoiding the additional burden of coordinate-level execution.
It therefore serves as the most basic behavioral probe of whether the model
can distinguish the sparse exceptions from the implicit majority reference.

\textbf{T2: Local counting (L-Cnt)} retains the same count output but introduces
a numerically specified row range, column range, or rectangle.
The model must now restrict the global exception set to the current region
before reporting its cardinality.
Because the required response remains \texttt{COUNT(n)}, the difference
between T1 and T2 primarily probes whether the visual evidence can be rebound
to a symbolic spatial context without yet requiring explicit localization
actions.

\paragraph{From contextual selection to exact action.}
The remaining tasks replace compact cardinality readout with explicit
coordinate-level execution.
\textbf{T3: Local clicking (L-Clk)} uses a numerically specified region, as in
T2, but requires the model to return the exact coordinates of all relevant
exceptions.
This transition makes it possible to compare context-conditioned counting with
context-conditioned action under closely related spatial instructions.

\textbf{T4: Visual-region clicking (V-Clk)} further changes how the permitted
region is specified.
Instead of receiving numeric row and column bounds, the model must ground a
region indicated directly in the image and then return the exception
coordinates within it.
This couples exact action with visual-region interpretation.

Finally, \textbf{T5: Exclusion clicking with count submission (Excl-CS)} asks
the model to select exception cells outside a specified region and to submit
the number of returned coordinates.
It therefore combines complement-based contextual filtering, exact coordinate
execution, and consistency between the selected action set and its reported
cardinality.

Taken together, T1 and T2 provide counting-oriented probes, while T3--T5 test
whether the same scene-level evidence supports increasingly explicit and
context-sensitive actions.
All five tasks are derived from the same underlying scene but are queried
independently in a single turn.
Their differences therefore measure cross-context behavioral consistency under
shared visual evidence rather than literal transfer of an internal model state.

Local regions are sampled to contain none, some, or all of the global
exceptions.
These cases respectively probe abstention, context-specific subset selection,
and complete-set execution.
Target placement, region sampling, and cue construction are detailed in
Appendix~\ref{app:scene_region_sampling}.

\subsection{Controlled Visual Sources and Dataset Composition}
\label{sec:source_curation}
\label{sec:dataset_stats}

ROSE uses multiple visual sources not merely to increase appearance diversity,
but to test whether the perception-to-action gap persists when the visual
evidence itself takes qualitatively different forms.
Each source provides a pair of majority and exception elements that can be
reused across the coupled task suite under matched rendering conditions.
The retained distinction must be fine-grained but human-visible, and the
exception identity is never named in the instruction.

The five subsets introduce complementary controls.
\textsc{ChineseGlyph} compares distinct but visually confusable characters
under the same verified font, while \textsc{EmojiStyle} holds semantic identity
fixed and changes only the rendering provider.
\textsc{EmojiContent} instead changes the depicted identity while using a
shared rendering style.
The two pixel-art subsets extend the benchmark beyond isolated symbolic icons:
\textsc{PixelEdit} introduces a localized change into the same source asset,
whereas \textsc{PixelContent} pairs distinct but visually related assets.

Here, \emph{pixel art} refers to the visual style of the source imagery rather
than to a fixed native resolution or a small-sprite setting.
The collected pool ranges from compact icons and individual objects to detailed
characters and complex scene-level compositions with substantial internal
structure.
This variation allows the same controlled task design to be tested on both
sparse symbolic forms and visually richer imagery.
Detailed source cleaning, pair construction, manual review, and representative
examples are provided in Appendix~\ref{app:visual_source_examples}.

\begin{table}[t]
\centering
\small
\setlength{\tabcolsep}{4.5pt}
\begin{tabular}{@{}llrrr@{}}
\toprule
\textbf{Subset} &
\textbf{Controlled visual source} &
\textbf{Scenes} &
\textbf{Dev} &
\textbf{Test} \\
\midrule

\textsc{ChineseGlyph}
& Confusable characters, same verified font
& 412 & 555 & 1505 \\

\textsc{EmojiStyle}
& Same emoji, different rendering providers
& 300 & 395 & 1105 \\

\textsc{EmojiContent}
& Related emoji identities, shared rendering style
& 300 & 395 & 1105 \\

\textsc{PixelEdit}
& Same pixel-art asset, localized edit
& 300 & 395 & 1105 \\

\textsc{PixelContent}
& Related but distinct pixel-art assets
& 200 & 260 & 740 \\
\midrule

\textbf{Total}
& {}
& 1512 & 2000 & 5560 \\
\bottomrule
\end{tabular}
\caption{
Controlled visual sources and dataset composition of ROSE v0.1.
Each scene produces five task instances; Dev and Test report task-instance
counts under the official scene-level split.
}
\label{tab:subset_curation}
\end{table}

ROSE v0.1 contains 1,512 scenes, 3,024 rendered images, and 7,560 task
instances.
Each scene produces five coupled tasks and two renderings: an uncued base image
and a cue-augmented image used for visual-region clicking.

The benchmark is split at the scene level.
All task variants and both renderings derived from the same scene are assigned
to the same split, preventing closely related variants of a test scene from
appearing during development.
The development split contains 2,000 instances for prompt and protocol
validation, while the test split contains 5,560 instances for the reported
evaluation.

\section{Experiments}

\subsection{Setup}
\label{sec:setup}

\paragraph{Models and inference.}
We evaluate the nine recent MLLMs listed in Table~\ref{tab:main_results}.
Each item is queried independently with a single image and no conversational history.
Temperature is set to 0, and explicit reasoning or thinking modes are disabled where supported.
Models are instructed to return only the formal answer, and one successful completion is retained and scored for each item.
Request-level retries are used only when no API response is returned.
Complete model identifiers and per-run inference configurations are released with the evaluation code.

\paragraph{Prompting.}
Each query consists of a shared protocol instruction, the grid image, and a task-specific instruction.
The shared instruction defines the row--column indexing convention, the allowed \texttt{COUNT}/\texttt{CLICK}/\texttt{SUBMIT} grammar, and format-only examples.
The task-specific instruction specifies the relevant region and required output operation.
No task-solving demonstrations are provided, and the complete prompt templates are included in the appendix and released code.

\paragraph{Evaluation metrics.}
We report PASS as the primary metric.
For counting tasks, PASS requires an exactly correct count in the prescribed grammar.
For click-based tasks, the predicted coordinate set must exactly match the ground-truth set; click order is ignored, whereas malformed outputs, out-of-grid coordinates, and repeated coordinates are treated as failures.
For click-and-submit tasks, the submitted count must additionally match both the ground-truth count and the number of unique predicted clicks.
VALID reports whether the response satisfies the task-specific output grammar, independently of whether the resulting answer is correct.

SOFT provides output-specific partial credit.
For counting tasks, it is defined as $1/(1+|\hat{n}-n|)$.
For click tasks, it is the coordinate-set F1 score, set to zero when the output is malformed, contains an out-of-grid coordinate, or repeats a coordinate.
For click-and-submit tasks, SOFT averages the strict coordinate F1, count SOFT, and an indicator of consistency between the submitted count and the number of unique predicted clicks.
C-F1 denotes the same strict coordinate-set F1 averaged over click-applicable tasks.
R-OK is the complement of the region-violation rate, measuring whether predicted in-grid clicks remain within the required region.
Unless otherwise stated, results are first computed within each visual subset and then macro-averaged equally over the five subsets.

\paragraph{Human reference.}
For reference, we evaluate a trained human annotator on the official test split through a custom web interface with access to image zooming.
The interface directly records numeric responses for counting tasks and the final set of selected grid cells for click-based tasks, rather than requiring the model-facing formal grammar.
Human responses are evaluated using the same task-level success criteria as model predictions: exact count match for counting tasks and exact coordinate-set match under the specified region condition for click-based tasks.
Because selected cells are stored as a set, repeated click actions are not represented in the submitted response.
The resulting score is used only as a solvability reference for ROSE.

\subsection{Main Results}

\begin{table*}[t]
\centering
\small
\setlength{\tabcolsep}{4.2pt}
\renewcommand{\arraystretch}{1.06}

\begin{tabular}{@{}l*{10}{c}@{}}
\toprule
\multirow{2}{*}{\textbf{Model}} &
\multicolumn{5}{c}{\textbf{Task PASS}} &
\multicolumn{3}{c}{\textbf{Visual-Source PASS}} &
\multirow{2}{*}{\textbf{Avg.}} &
\multirow{2}{*}{\textbf{VALID}} \\
\cmidrule(lr){2-6}
\cmidrule(lr){7-9}
&
\textbf{G-Cnt} &
\textbf{L-Cnt} &
\textbf{L-Clk} &
\textbf{V-Clk} &
\textbf{Excl-CS} &
\textbf{Glyph} &
\textbf{Emoji} &
\textbf{Pixel} &
&
\\
\midrule

Qwen3-VL-Flash
& 47.7 & 21.6 & 1.3 & 0.5 & 0.7
& 15.0 & 14.9 & 13.6
& 14.3 & 86.6 \\

Qwen3-VL-Plus
& 66.4 & 30.6 & 4.1 & 5.7 & 3.2
& 21.8 & 24.0 & 20.1
& 22.0 & 95.5 \\

Qwen3.6-Plus
& 80.3 & 65.3 & 39.5 & 37.7 & 28.9
& 48.4 & 48.0 & 53.6
& 50.3 & 99.9 \\

Claude-Sonnet-4.6
& 62.1 & 21.6 & 9.8 & 20.6 & 4.5
& 28.5 & 25.0 & 20.1
& 23.7 & 61.3 \\

Claude-Opus-4.8
& 64.0 & 21.2 & 9.8 & 21.4 & 4.9
& 30.2 & 25.2 & 20.4
& 24.3 & 62.7 \\

GLM-4.6V
& 60.7 & 30.8 & 5.0 & 4.5 & 2.5
& 19.1 & 22.4 & 19.9
& 20.7 & 98.8 \\

GLM-5V-Turbo
& 64.2 & 56.9 & 20.6 & 21.4 & 6.1
& 37.5 & 34.5 & 31.3
& 33.8 & 99.5 \\

Gemini-3.1-Pro
& 92.8 & 93.9 & 75.4 & 64.2 & 70.4
& 67.6 & 84.5 & 80.1
& 79.4 & 93.4 \\

GPT-5.5
& 93.8 & 97.0 & 93.6 & 84.3 & 92.5
& 87.4 & 94.8 & 92.2
& 92.2 & 100.0 \\

\midrule

Human
& 99.9 & 100.0 & 98.8 & 97.7 & 95.8
& 99.8 & 97.5 & 99.8
& 98.8 & -- \\

\bottomrule
\end{tabular}

\caption{
Primary ROSE results on the test split.
Task columns report strict PASS for the five coupled templates.
Glyph denotes \textsc{ChineseGlyph}; Emoji averages
\textsc{EmojiStyle} and \textsc{EmojiContent}; Pixel averages
\textsc{PixelEdit} and \textsc{PixelContent}.
Avg.\ remains the equal macro average over the five original visual subsets,
and VALID reports the grammar-valid output rate.
Full subset-level PASS and SOFT results are provided in the appendix.
}
\label{tab:main_results}
\end{table*}

\begin{figure}[t]
\centering
\includegraphics[width=0.9\linewidth]{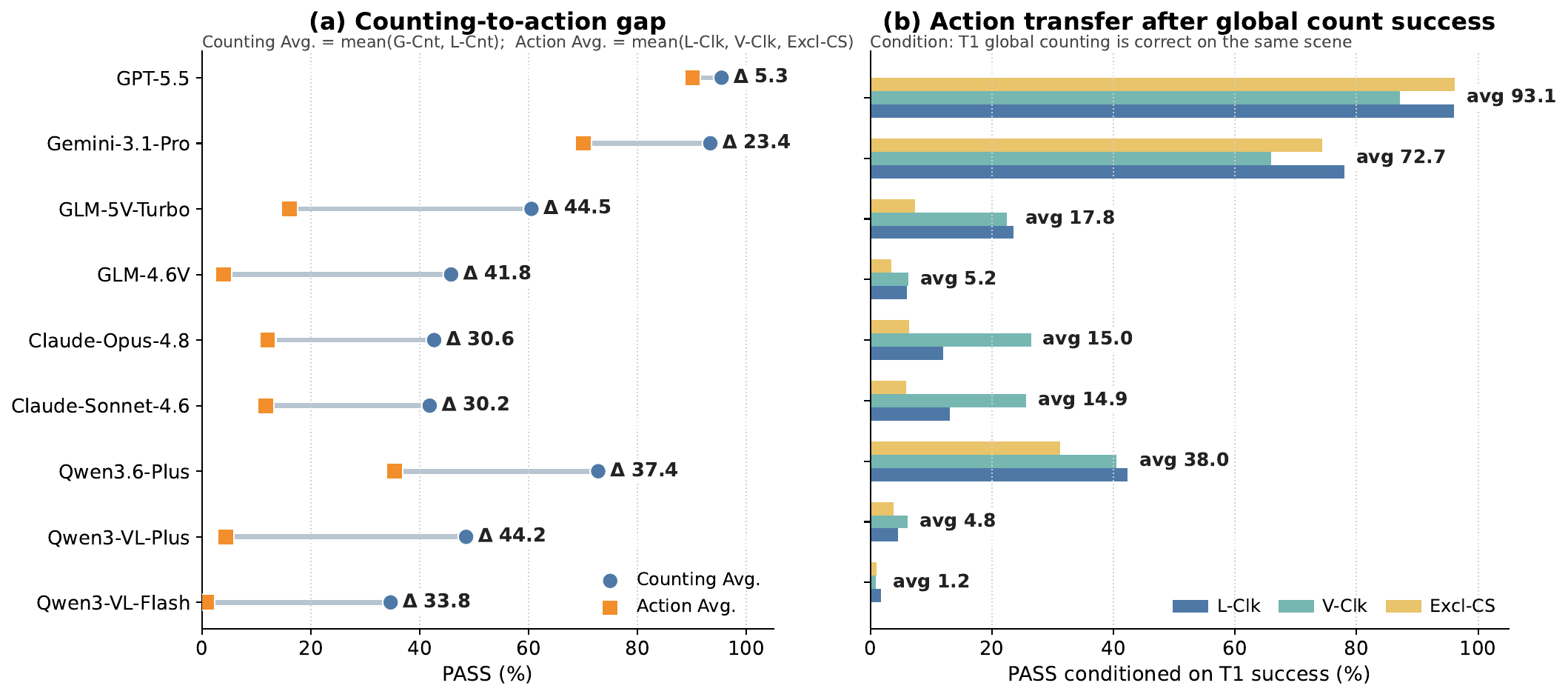}
\caption{
The primary counting-to-action gap in ROSE.
(a) Counting Avg.\ averages G-Cnt and L-Cnt, while Action Avg.\ averages
L-Clk, V-Clk, and Excl-CS.
(b) Each action task is evaluated only on paired scenes where the same model's
independent G-Cnt response is correct.
}
\label{fig:perception_action_gap}
\end{figure}

\paragraph{Current MLLMs often see enough to count, but not enough to act.}
Table~\ref{tab:main_results} shows that ROSE is highly solvable yet remains
strongly discriminative: the human reference reaches 98.8\% average PASS,
whereas model performance ranges from 14.3\% to 92.2\%.
More importantly, the dominant weakness is not confined to one visual source
or one model family, but appears when a compact visual readout must be converted
into exact, context-dependent action.
As shown in Figure~\ref{fig:perception_action_gap}(a), Qwen3.6-Plus drops from
72.8\% Counting Avg.\ to 35.3\% Action Avg., GLM-5V-Turbo from 60.5\% to
16.0\%, and Gemini-3.1-Pro from 93.4\% to 70.0\%.
GPT-5.5 narrows this gap substantially, from 95.4\% to 90.1\%, showing that
the drop is not an unavoidable consequence of the formal action protocol.
Nor is it explained simply by scenes that the model fails to interpret at all:
Figure~\ref{fig:perception_action_gap}(b) evaluates action only on paired scenes
where the same model independently answers global counting correctly, yet
conditioned Action Avg.\ remains only 38.0\% for Qwen3.6-Plus and 17.8\% for
GLM-5V-Turbo.
The primary result of ROSE is therefore not merely that clicking is harder than
counting, but that many models fail to preserve the usefulness of the same
visual evidence once the required output becomes region-sensitive and
coordinate-exact.

\subsection{Analysis}
\label{sec:analysis}

\subsubsection{Decomposing the Perception-to-Action Gap}
\label{sec:gap_decomposition}

\begin{table}
\centering
\small
\setlength{\tabcolsep}{3.2pt}
\renewcommand{\arraystretch}{1.08}

\begin{tabular*}{\linewidth}{
@{\extracolsep{\fill}}
l
cccc
ccc
cc
@{}
}
\toprule
&
\multicolumn{4}{c}{\textbf{Task PASS}} &
\multicolumn{3}{c}{\textbf{Bridge Diagnostics}} &
\multicolumn{2}{c}{\textbf{Performance Drop}} \\
\cmidrule(lr){2-5}
\cmidrule(lr){6-8}
\cmidrule(lr){9-10}

\textbf{Model} &
\textbf{G-Cnt} &
\textbf{G-Clk} &
\textbf{G-Clk}$^{\dagger}$ &
\textbf{V-Clk} &
\textbf{Card} &
\textbf{Loc.$\mid$Card} &
\textbf{C-F1} &
$\boldsymbol{\Delta}_{\mathrm{C\rightarrow G}}$ &
$\boldsymbol{\Delta}_{\mathrm{G\rightarrow V}}$ \\
\midrule

Qwen3.6-Plus
& 80.3 & 67.1 & 73.1 & 37.7
& 86.0 & 78.0 & 76.4
& -13.2 & -29.4 \\

Gemini-3.1-Pro
& 92.8 & 86.5 & 90.5 & 64.2
& 91.2 & 94.8 & 89.7
& -6.3 & -22.3 \\

GPT-5.5
& 93.8 & 91.8 & 96.1 & 84.3
& 94.9 & 96.7 & 95.4
& -2.0 & -7.5 \\

\bottomrule
\end{tabular*}

\caption{
Global-click bridge analysis.
G-Clk$^{\dagger}$ denotes global-click PASS restricted to scenes
where global counting is correct.
Card measures exact clicked cardinality, while
Loc.$\mid$Card measures exact coordinate localization conditioned
on correct cardinality.
$\Delta_{\mathrm{C\rightarrow G}}
=\mathrm{G\text{-}Clk}-\mathrm{G\text{-}Cnt}$ and
$\Delta_{\mathrm{G\rightarrow V}}
=\mathrm{V\text{-}Clk}-\mathrm{G\text{-}Clk}$.
}
\label{tab:global_click_bridge}
\end{table}

\paragraph{Does the action gap reduce to coordinate grounding?}
The preceding results show a clear gap between global counting and
region-conditioned clicking, but correct counting alone does not guarantee that
a model has localized every odd cell correctly.
To separate coordinate grounding from context-conditioned selection, we
introduce a global-click bridge task on three representative models.
Given the same uncued scene as global counting, the model is asked to click all
odd cells in the full grid, without any region restriction.
This yields a three-stage comparison from global counting
(G-Cnt), to global coordinate localization (G-Clk), and finally to
visual-region clicking (V-Clk).

Table~\ref{tab:global_click_bridge} shows a consistent ordering across all
three models: performance decreases from G-Cnt to G-Clk and decreases further
from G-Clk to V-Clk.
For Qwen3.6-Plus, performance drops from 80.3\% G-Cnt to 67.1\% G-Clk,
indicating a measurable count-to-coordinate gap, but the larger drop occurs
after introducing the region context, with V-Clk falling to 37.7\%.
Gemini-3.1-Pro exhibits the same pattern, decreasing by 6.3 points from
G-Cnt to G-Clk but by a further 22.3 points from G-Clk to V-Clk.
GPT-5.5 nearly preserves its global-counting performance under global
clicking (93.8\% versus 91.8\%), yet still declines to 84.3\% once
region-conditioned selection is required.
Therefore, coordinate grounding explains only part of the original
perception-to-action gap; for all three models, the additional loss introduced
by the region context is larger than the preceding count-to-coordinate loss.

The conditioned results further support this conclusion.
When evaluation is restricted to scenes where the same model answers global
counting correctly, G-Clk remains 73.1\%, 90.5\%, and 96.1\% for
Qwen3.6-Plus, Gemini-3.1-Pro, and GPT-5.5, respectively.
Thus, even after conditioning on correct global-count responses,
Qwen3.6-Plus and Gemini-3.1-Pro retain substantially stronger global
localization than region-conditioned action. 
The diagnostics also reveal different intermediate bottlenecks:
Qwen3.6-Plus obtains 86.0\% exact clicked cardinality but only 78.0\%
exact localization when cardinality is correct, showing that part of its loss
comes from selecting the wrong coordinates.
In contrast, Gemini-3.1-Pro and GPT-5.5 reach 94.8\% and 96.7\%
localization accuracy under correct cardinality, suggesting that their remaining
failures are less attributable to basic coordinate mapping.
Overall, the bridge analysis decomposes the action gap into two distinct
components---converting a visual count into exact coordinates and rebinding
those coordinates to the current task context---with the latter forming the
larger bottleneck.

\begin{table}[t]
\centering

\begin{minipage}{0.62\linewidth}
\centering
\footnotesize
\setlength{\tabcolsep}{3.2pt}
\renewcommand{\arraystretch}{1.06}

\resizebox{\linewidth}{!}{%
\begin{tabular}{@{}llcccc@{}}
\toprule
\multirow{2}{*}{\textbf{Model}} &
\multirow{2}{*}{\textbf{Case}} &
\multicolumn{2}{c}{\textbf{Unconditioned PASS}} &
\multicolumn{2}{c}{\textbf{Matched Consistency}} \\
\cmidrule(lr){3-4}
\cmidrule(lr){5-6}
&
&
\textbf{mL-Cnt} &
\textbf{L-Clk} &
\textbf{L-Clk}$^{\dagger}$ &
\textbf{Fail}$^{\dagger}$ \\
\midrule

\multirow{4}{*}{Qwen3.6-Plus}
& Overall & 63.2 & 39.5 & 52.7 & 47.3 \\
& Zero    & 67.5 & 12.7 & 18.3 & 81.7 \\
& Partial & 55.8 & 47.3 & 67.7 & 32.3 \\
& All     & 84.7 & 73.8 & 78.7 & 21.3 \\

\midrule

\multirow{4}{*}{GPT-5.5}
& Overall & 96.7 & 93.6 & 95.8 & 4.2 \\
& Zero    & 99.2 & 92.8 & 93.6 & 6.4 \\
& Partial & 95.5 & 93.7 & 96.8 & 3.2 \\
& All     & 94.0 & 91.0 & 94.0 & 6.0 \\

\bottomrule
\end{tabular}%
}

\caption{
Matched local count-to-click consistency.
Each mL-Cnt and L-Clk pair uses the same image, numeric region, and regional
target set.
L-Clk$^{\dagger}$ denotes local-click PASS conditioned on correct independently
queried matched local counting, and
Fail$^{\dagger}=100-\mathrm{L\text{-}Clk}^{\dagger}$.
Zero, Partial, and All are defined by the realized relation between the regional
and global target sets; fallback generation cases are merged into the
corresponding realized category.
Results are macro-averaged over the five visual subsets.
}
\label{tab:matched_local_transfer}

\end{minipage}
\end{table}

\paragraph{Does correct regional counting support exact action?}
The original L-Cnt and L-Clk templates use independently sampled regions, so
their aggregate difference may partly reflect variation in region difficulty.
We therefore construct an exactly matched control in which the image, numeric
region, and regional target set are held fixed, while only the required output
changes from \texttt{COUNT} to coordinate-level \texttt{CLICK} actions.

Table~\ref{tab:matched_local_transfer} reveals a sharp model-dependent
difference.
For Qwen3.6-Plus, local-click accuracy reaches only 52.7\% among matched
regions that it independently counts correctly, leaving a 47.3\% conditional
failure rate.
The failure is particularly severe for Zero regions: even after returning the
correct count of zero, the model produces the required empty action in only
18.3\% of paired queries.
Partial and All regions improve to 67.7\% and 78.7\%, but correct cardinality
still frequently fails to coincide with the exact coordinate set.

GPT-5.5 behaves very differently.
It reaches 96.7\% on matched local counting and 93.6\% on the paired local-click
tasks.
Conditioned on correct matched counting, exact local-click accuracy remains
95.8\% overall and ranges from 93.6\% to 96.8\% across Zero, Partial, and All
regions.
Thus, correct regional cardinality is highly predictive of exact action for
GPT-5.5, although a small residual inconsistency remains, especially when the
required action is empty or includes the complete regional target set.

The matched control therefore shows that the perception-to-action gap is not a
fixed consequence of changing the output grammar.
Some models, exemplified by Qwen3.6-Plus, frequently fail to produce the exact
regional action even when they independently recover the correct cardinality
under an otherwise identical context.
Stronger models can largely close this gap, indicating that cross-task
consistency is a substantive and model-dependent capability rather than an
unavoidable property of coordinate output.

\subsubsection{Action Failure Modes}
\label{sec:failure_modes}

\begin{figure*}[t]
\centering
\includegraphics[width=\textwidth]{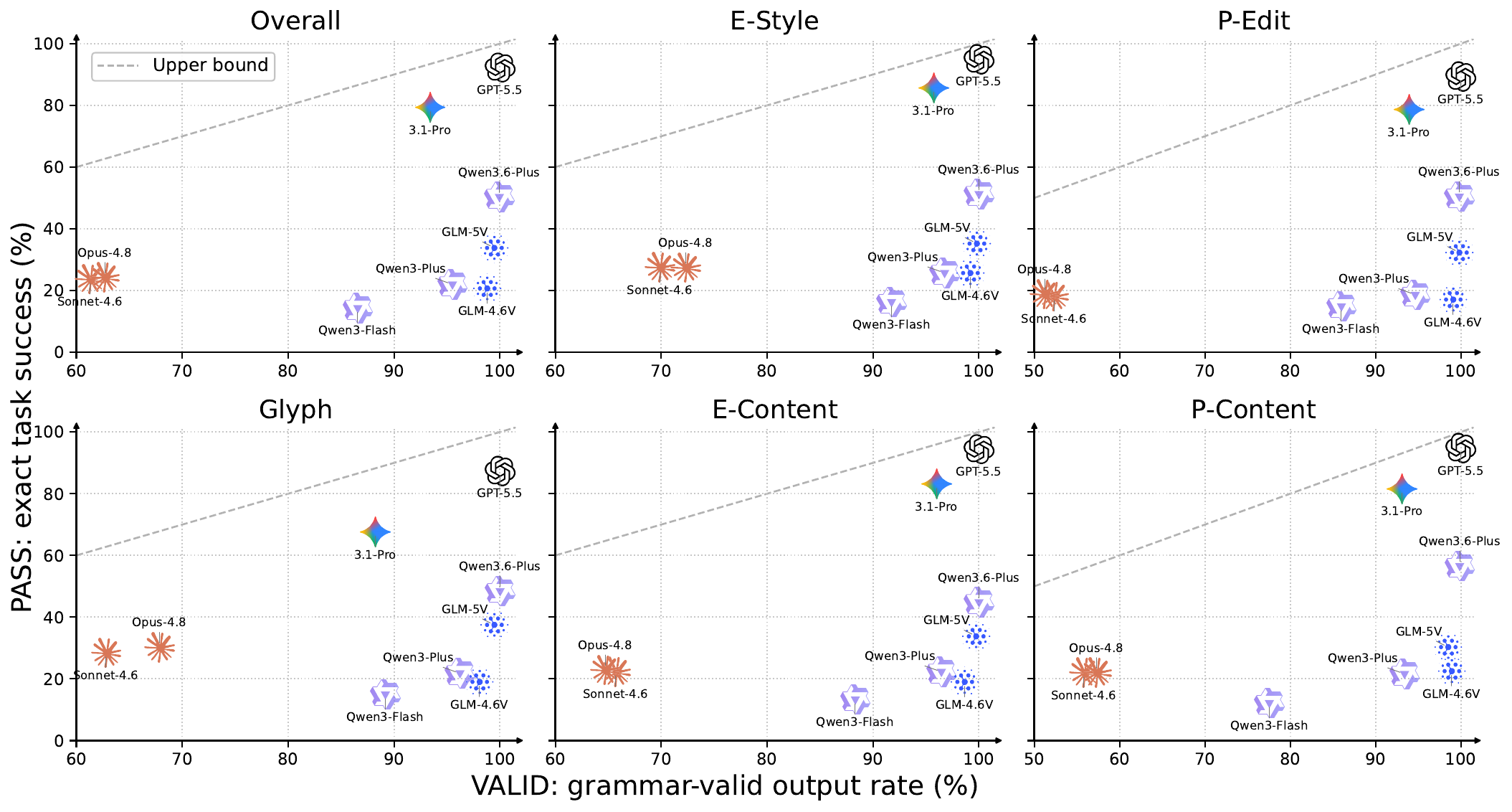}
\caption{
VALID--PASS decoupling across ROSE subsets.
The dashed line denotes the validity-limited upper bound: a prediction cannot receive PASS unless it first satisfies the formal output grammar.
}
\label{fig:valid_pass}
\end{figure*}

\paragraph{Are failures mainly caused by invalid output format?}
Figure~\ref{fig:valid_pass} separates formal output validity from actual task success.
Because a strict PASS requires a grammar-valid answer, the dashed line is not a performance trend but a validity-limited upper bound: models near this line mostly fail because they are invalid, while models far below it produce formally valid answers that are nevertheless wrong.
The overall panel shows that several models fall into the latter regime.
Qwen3.6-Plus, GLM-4.6V, and GLM-5V-Turbo all achieve near-perfect VALID, yet remain far below their validity ceiling in PASS.
This means that their errors are not primarily caused by malformed \texttt{COUNT}/\texttt{CLICK}/\texttt{SUBMIT} strings, but by selecting the wrong cells, applying the wrong region condition, or converting the visual decision into an incorrect coordinate action.

The subset panels show that this decoupling is not an artifact of a single visual source.
Across Glyph, E-Style, E-Content, P-Edit, and P-Content, high-validity models such as Qwen and GLM consistently occupy the high-VALID but low-PASS region.
In contrast, the Claude models form a different failure regime, with much lower VALID and correspondingly limited PASS, indicating that protocol following is a substantial bottleneck for them.
Together, these patterns justify treating VALID as a control rather than as an explanation for low performance: ROSE exposes many failures that occur after the model has already produced a syntactically acceptable action.

\begin{figure*}[t]
\centering
\includegraphics[width=0.75\textwidth]{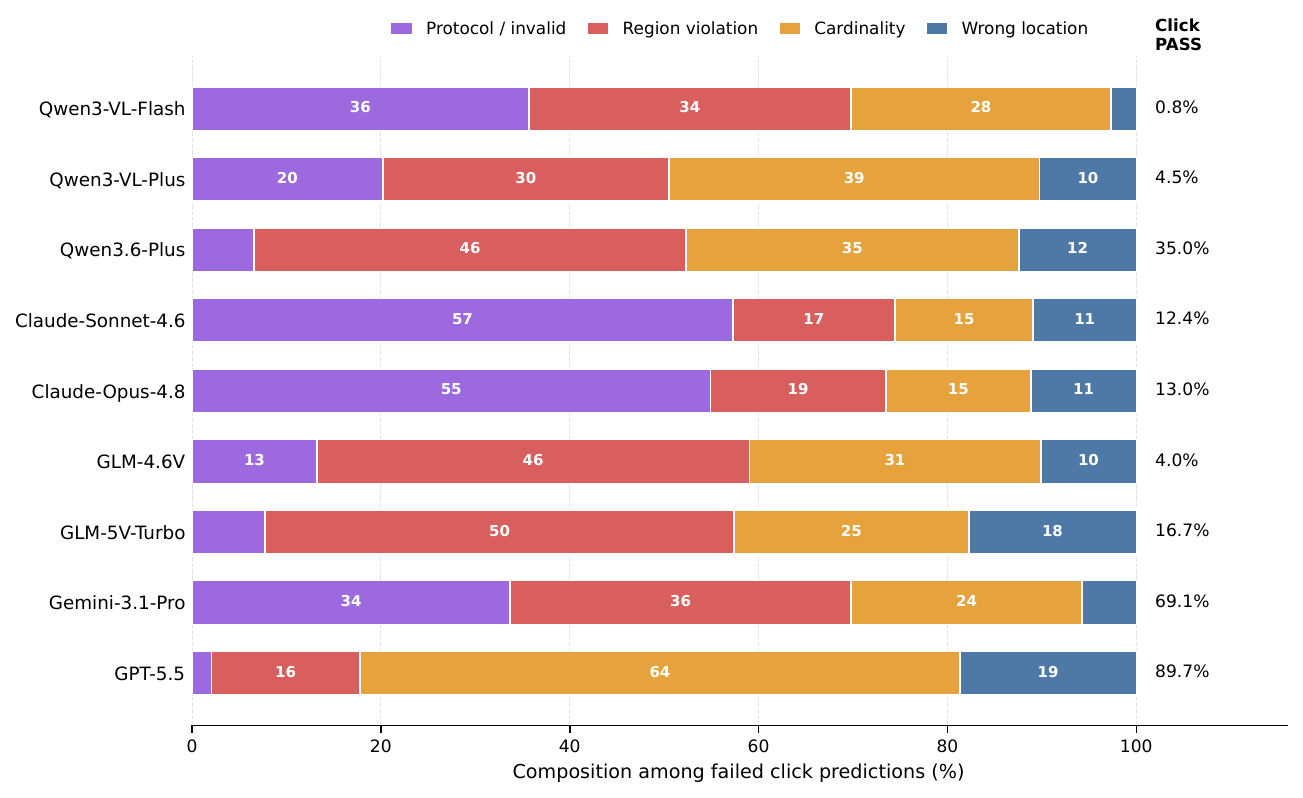}
\caption{
Click error typology on ROSE action tasks.
Bars show the composition of failed click-applicable predictions, normalized within each model's failed cases.
Protocol/invalid groups malformed outputs, invalid coordinates, and click--submit inconsistency.
The right column reports the overall Click PASS rate.
}
\label{fig:click_error_typology}
\end{figure*}

\paragraph{How do action predictions fail?}
Figure~\ref{fig:click_error_typology} decomposes failed click-applicable predictions into coarse, mutually exclusive error types.
Unlike Figure~\ref{fig:valid_pass}, which separates valid from invalid outputs, this analysis asks what remains after a model attempts an executable action.
The resulting failure profiles differ substantially across model families.
The Claude models are dominated by protocol/invalid errors, accounting for 57\% and 55\% of their failed click predictions, indicating that executable action formatting is a major bottleneck for them.
In contrast, GLM-4.6V and GLM-5V-Turbo are dominated by region violations, while Qwen3.6-Plus also shows a large region-violation component.
These errors are especially diagnostic for ROSE: the model often produces a syntactically valid action, but applies it under the wrong region constraint.

The stronger models exhibit a different pattern.
Gemini-3.1-Pro achieves a much higher Click PASS rate, but its remaining failures are still split across protocol, region, and cardinality errors.
GPT-5.5 reaches 89.7\% Click PASS, and its few remaining errors are mostly cardinality errors rather than malformed outputs or broad region violations.
This suggests that as models become stronger, the dominant bottleneck shifts from producing valid actions and respecting context constraints toward the exact execution of the required number of coordinate-level clicks.
Overall, ROSE not only exposes a perception-to-action gap, but also separates the failure modes behind that gap.

\subsubsection{Qualitative Failure Cases}
\label{sec:qualitative}

\begin{figure*}[t]
\centering
\includegraphics[width=\textwidth]{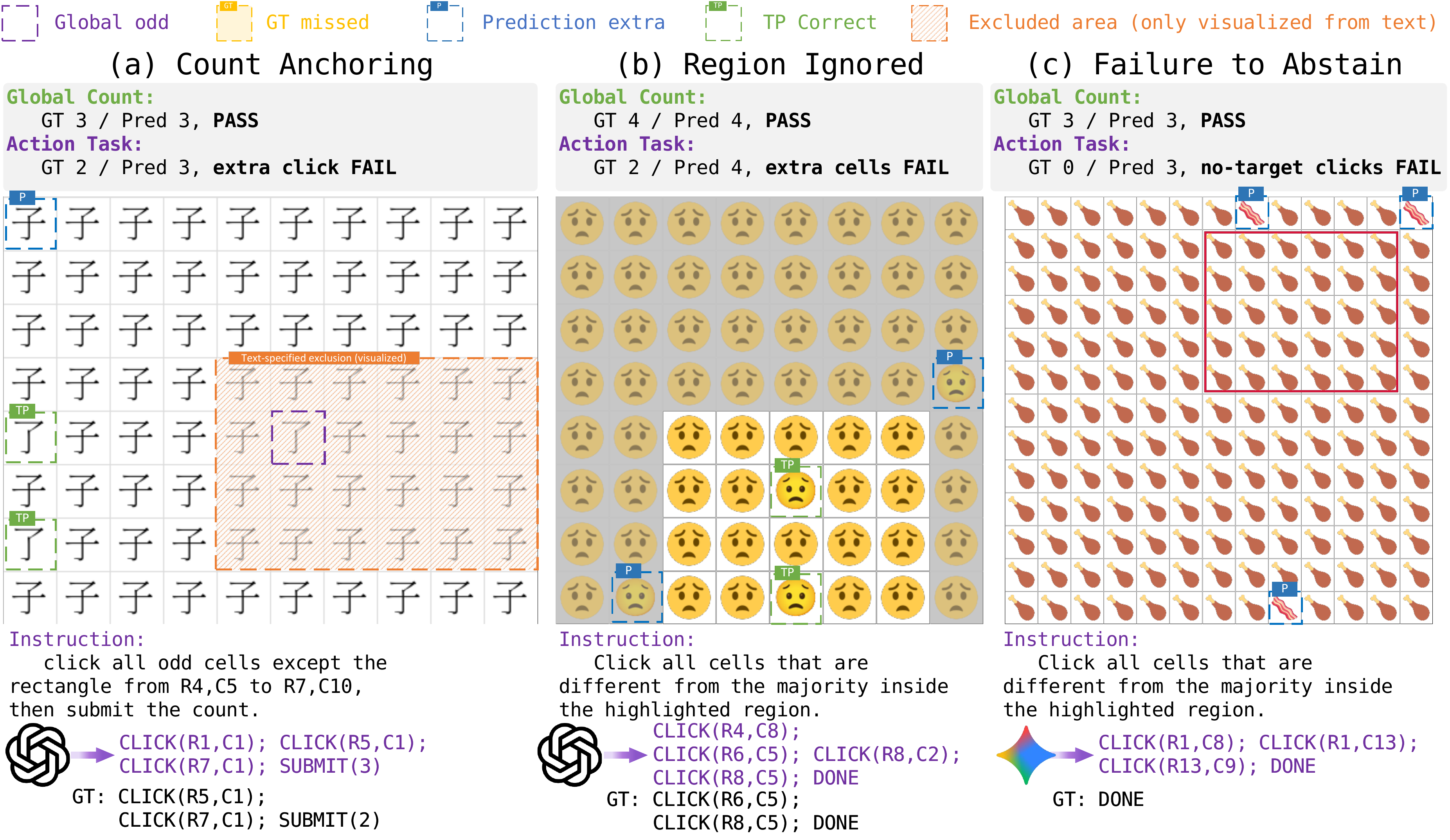}
\caption{
Qualitative failure cases in ROSE.
In each example, the model's independently queried global-count response is
correct, but its region-conditioned action is not.
Purple dashed boxes mark global exception cells, green boxes mark correct
clicks, blue boxes mark prediction-only extra clicks, and orange hatched
regions visualize text-specified exclusion areas.
The examples illustrate count anchoring, failure to apply the current region,
and failure to abstain when no valid target remains.
}
\label{fig:qualitative_failures}
\end{figure*}

Figure~\ref{fig:qualitative_failures} visualizes representative failures paired
with correct global-count responses.
These cases show that recovering the global exception cardinality does not
ensure that the same visual evidence will support the exact action required by
the current context.
Instead, models may retain the global answer or target set even after the task
condition changes which cells are actionable.

In Figure~\ref{fig:qualitative_failures}(a), the model correctly reports three
global exceptions, but the exclusion condition removes one of them from the
valid action set.
The correct response therefore contains two clicks and
\texttt{SUBMIT(2)}.
The model instead returns three clicks and preserves the original global
cardinality.
This \emph{count anchoring} failure indicates that the exclusion condition is
not fully reflected in the recomputed action set.

Figure~\ref{fig:qualitative_failures}(b) shows a more direct region-filtering
failure.
Although four exception cells exist globally, only two lie inside the
highlighted region.
The prediction includes both valid targets but also clicks additional global
exceptions outside the permitted region, effectively reverting to the
full-scene exception set rather than selecting its context-relevant subset.

Finally, Figure~\ref{fig:qualitative_failures}(c) illustrates failure to
abstain.
The highlighted region contains no exception cell, so the correct response is
simply \texttt{DONE}.
Nevertheless, the model returns three global exception coordinates.
Together, these cases expose three closely related failures at the
perception-to-action interface: updating the target cardinality, filtering the
global target set under the current region, and suppressing action when the
resulting target set is empty.

\section{Conclusion}
\label{sec:conclusion}

We introduced \textsc{ROSE}, a controlled benchmark for evaluating whether
multimodal models can convert fine-grained visual evidence into
context-dependent symbolic actions.
By coupling counting and coordinate-action tasks over the same underlying
scene, \textsc{ROSE} makes it possible to examine how consistently shared
visual evidence supports different region constraints and output operations.
Experiments across recent MLLMs reveal a clear and strongly model-dependent
counting-to-action gap, despite near-ceiling human performance.
This gap cannot be explained by malformed outputs or coordinate grounding
alone: even when global or regional cardinality is correct, some models still
fail to produce the exact action set required by the current context.
These failures commonly appear as region violations, incorrect cardinality
updates, and failure to abstain when no valid target remains.
Together, the results suggest that reliable multimodal action requires not only
recovering relevant visual evidence, but also recomputing the
context-specific target set and executing it exactly.
We hope \textsc{ROSE} provides a useful controlled testbed for measuring
progress toward this capability.

\section*{Limitations}

ROSE is intentionally designed as a controlled diagnostic benchmark rather than a comprehensive simulation of open-world visual interaction. Its grid-structured scenes and symbolic action space make perception, region selection, and execution directly measurable, but do not capture the full complexity of natural images, free-form language, long-horizon planning, or environment feedback. The current version covers five visual sources and a fixed set of counting and clicking templates; broader visual domains, richer reference relations, and interactive or temporal settings remain important directions for extension. In addition, the evaluated model scores reflect specific model versions and inference configurations, and should therefore be interpreted as diagnostic evidence rather than a permanent ranking. The human result is used primarily as a solvability reference, rather than as a population-level estimate of human performance.

\section*{Broader Impact}

ROSE provides a controlled way to study whether multimodal models can turn fine-grained visual evidence into context-sensitive, executable decisions. Such capabilities are relevant to reliable visual agents, GUI interaction, document processing, assistive systems, and other applications where selecting the correct visual target is not sufficient unless the model also respects the current task constraint. By releasing scene-coupled tasks, exact evaluators, and diagnostic analyses, ROSE can support reproducible comparison and help identify whether failures arise from perception, region binding, coordinate grounding, or action execution. At the same time, strong performance on ROSE should not be interpreted as a guarantee of safety or competence in unconstrained real-world environments, where semantic ambiguity, dynamic feedback, and broader operational risks remain.

\bibliography{iclr2026_conference}
\bibliographystyle{iclr2026_conference}

\appendix

\section{Visual Source Construction and Quality Control}
\label{app:visual_source_examples}

This section provides the implementation details behind the visual-source and
scene-construction controls summarized in Section~\ref{sec:benchmark}.
ROSE prioritizes controlled, fine-grained, and human-visible differences rather
than unrestricted visual diversity.
Across all subsets, a retained visual pair should satisfy three basic
requirements: the difference should be sufficiently subtle to require direct
visual comparison, sufficiently clear to remain human-solvable, and rendered
under matched conditions so that the majority--exception relation cannot be
resolved through an unintended source or formatting cue.

The five subsets instantiate these requirements at different levels of visual
variation.
\textsc{ChineseGlyph} uses distinct but visually confusable written symbols;
\textsc{EmojiStyle} varies rendering style while preserving emoji identity;
\textsc{EmojiContent} varies depicted content under a shared rendering style;
\textsc{PixelEdit} introduces a localized modification to the same source
image; and \textsc{PixelContent} pairs distinct but visually related pixel-art
assets.
After source curation, all retained pairs are passed through the same
scene-generation, region-sampling, and task-instantiation pipeline.

\subsection{Chinese Glyph Pair Cleaning and Font Verification}
\label{app:glyph_examples}

\paragraph{Character-group cleaning.}
The \textsc{ChineseGlyph} source pool begins with manually collected groups of
visually confusable Chinese characters.
Because the raw groups originate from heterogeneous records, they may contain
mixed separators, repeated characters, invisible Unicode marks, or partially
duplicated groups.
We normalize the raw entries by removing invisible marks, splitting mixed
separators, eliminating repeated tokens, and discarding groups that are strict
subsets of already retained groups.

Each cleaned group is expanded into undirected character pairs.
Pairs are deduplicated without regard to direction, so that the relation between
two characters is not counted twice merely by reversing their order.
Each retained pair is used once in the final source pool, with one character
assigned as the majority element and the other as the exception element during
scene construction.
This preserves a broad range of glyph-level confusions while limiting repeated
use of the same character relation.

\paragraph{Font verification.}
Chinese glyph rendering requires additional control because nominal font
support does not guarantee a usable visual rendering.
Some fonts lack one or both characters in a pair, replace unsupported
characters with missing-glyph boxes, or produce unstable and unreadable forms.
We therefore construct a verified font pool by screening candidate fonts for
character coverage and rendering quality.

Within every generated scene, the majority and exception characters are
rendered using the same font, size, and rendering configuration.
The model therefore cannot identify the exception through a change in typeface
or rasterization settings.
Instead, it must compare the actual glyph structures, such as stroke position,
enclosure shape, internal strokes, or auxiliary marks.
A retained character pair may appear more or less similar under different
fonts, but the comparison remains within-font for every individual scene.

\begin{figure*}[t]
\centering
\includegraphics[width=\textwidth]{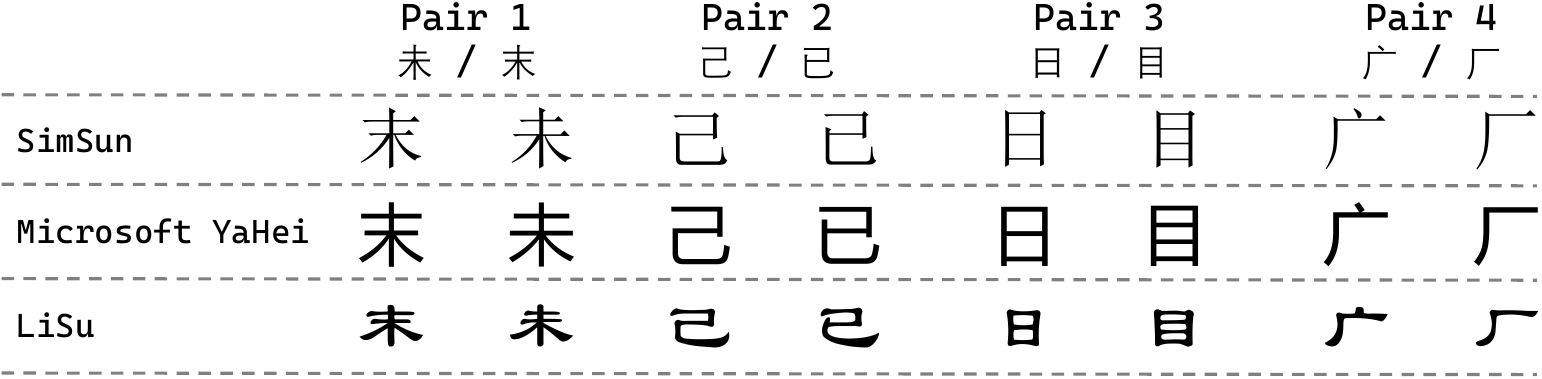}
\caption{
Representative Chinese glyph pairs rendered under three verified fonts.
Each column denotes one retained pair, and each row renders all pairs using
the same typeface.
The examples cover subtle differences in stroke position, enclosure structure,
internal strokes, and auxiliary marks.
Each pair contains two distinct Chinese characters rather than two renderings
of the same character.
}
\label{fig:app_glyph}
\end{figure*}

Figure~\ref{fig:app_glyph} shows representative retained pairs under several
verified fonts.
The examples also illustrate why font verification is necessary: the location
and visual prominence of the distinguishing stroke may change across
typefaces, even though the two characters remain distinct and human-readable.

\subsection{Emoji Asset Filtering and Pair Selection}
\label{app:emoji_examples}

\paragraph{Asset collection and filtering.}
For the emoji subsets, we collect public emoji metadata together with rendering
assets from multiple providers.
The metadata records include the emoji character, its name, and its semantic
category.
We exclude complex sequences whose rendering or asset availability is likely
to vary across providers, including flags, keycaps, tag sequences, long
zero-width-joiner chains, and overlong compound emoji.
This filtering reduces ambiguity caused by provider-dependent sequence
composition rather than by the intended style or content difference.

\paragraph{\textsc{EmojiStyle}.}
The \textsc{EmojiStyle} subset holds semantic identity fixed while changing the
rendering source.
For each retained identity, the majority and exception elements depict the
same emoji but use different provider-specific visual styles.
The resulting distinction may involve shape simplification, shading, outline
thickness, color distribution, facial details, or other rendering conventions.
Because the underlying emoji identity is unchanged, semantic recognition alone
is insufficient: the model must distinguish two visual realizations of the
same content.

Candidate identities are retained only when the required provider assets are
available and the resulting style difference is visually discernible.
Complex or inconsistently rendered sequences are removed before pair
construction.

\paragraph{\textsc{EmojiContent}.}
The \textsc{EmojiContent} subset instead keeps rendering style fixed while
changing emoji identity.
Candidate pairs are drawn from the same semantic category, reducing the
likelihood that the exception is an obviously unrelated object.
We render candidate emoji using a shared style and compute coarse visual
similarity features based on color and edge information.
Pairs are ranked and filtered using these similarities so that retained
elements remain visually related without becoming indistinguishable.

The selected pairs are additionally balanced across emoji categories.
This prevents the subset from being dominated by a small number of common
faces, gestures, or object types and reduces repeated semantic or visual
patterns.

\begin{figure*}[t]
\centering
\includegraphics[width=\textwidth]{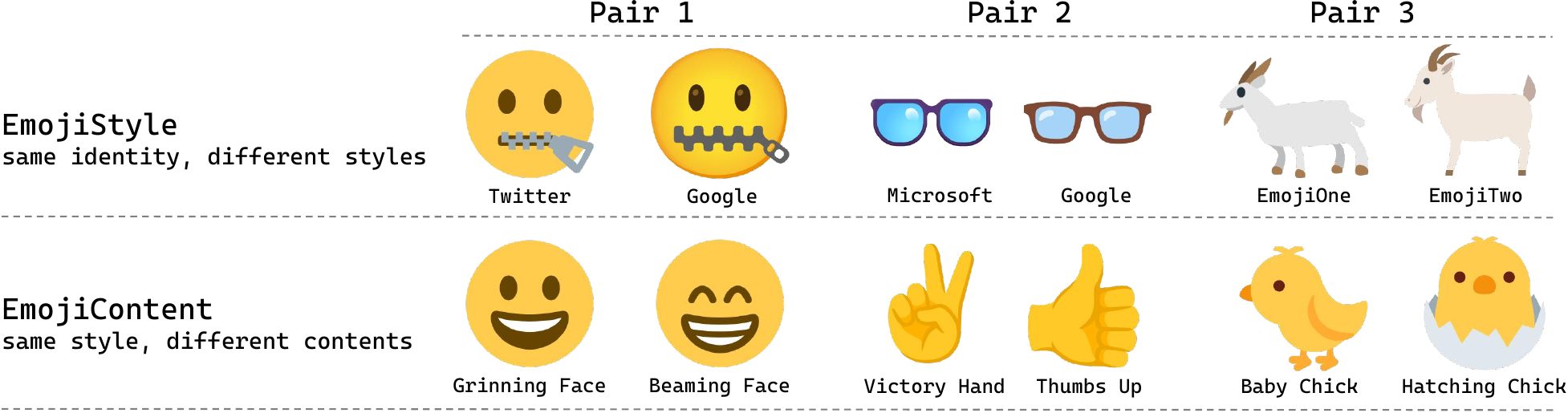}
\caption{ Representative pairs from \textsc{EmojiStyle} and \textsc{EmojiContent}. The first row shows the same emoji identities rendered by different providers, whose names are indicated below the corresponding elements. The second row shows visually related but distinct emoji contents rendered in a shared style, with their emoji names provided for reference. }
\label{fig:app_emoji}
\end{figure*}

Figure~\ref{fig:app_emoji} contrasts the two pairing strategies. The first row shows \textsc{EmojiStyle} pairs, where emoji identity is fixed but provider-specific rendering changes. The second row shows \textsc{EmojiContent} pairs, where rendering style is shared but the depicted identities differ.

\subsection{Pixel-Art Source Collection and Pair Construction}
\label{app:pixel_examples}

The pixel-art subsets are constructed from a pool of web-collected pixel
images with diverse subjects, resolutions, palettes, and visual styles.
The two subsets use this source pool differently:
\textsc{PixelEdit} isolates a localized visual modification within the same
source image, whereas \textsc{PixelContent} pairs two distinct but visually
related source assets.

\paragraph{\textsc{PixelEdit}.}
For each candidate source image, an edited variant is generated by introducing
a localized change while preserving the overall identity and composition of
the image.
The intended difference may affect a small object part, boundary, internal
pattern, accessory, color region, or other local structure.

Generated edits are manually reviewed before inclusion.
We remove candidates when the modified area is too subtle to identify
reliably, when the change is so disruptive that the pair becomes trivial, when
unintended text or global artifacts are introduced, or when the before--after
relation is otherwise ambiguous.
Only pairs with a visible but localized distinction are retained in the source
pool used by the benchmark generator.

This review is important for interpreting failures on \textsc{PixelEdit}.
The subset is intended to test sensitivity to local visual changes, rather than
whether a model can detect an edit that is effectively invisible even under
careful human inspection.

\paragraph{\textsc{PixelContent}.}
\textsc{PixelContent} uses two independently collected assets rather than an
edited source--variant pair.
To avoid arbitrary pairings, we tokenize available filenames and infer coarse
themes from shared name tokens.
Candidate pairs are generated when the assets share a token, subject, or broad
theme.

These candidates are further filtered using coarse visual features, including
color histograms, edge statistics, and downsampled image representations.
The filtering favors pairs that are visually related in palette or structure
while still depicting distinct contents.
We also cap repeated asset usage so that a small number of generic or visually
common images cannot dominate the subset.

\begin{figure*}[t]
\centering
\includegraphics[width=\textwidth]{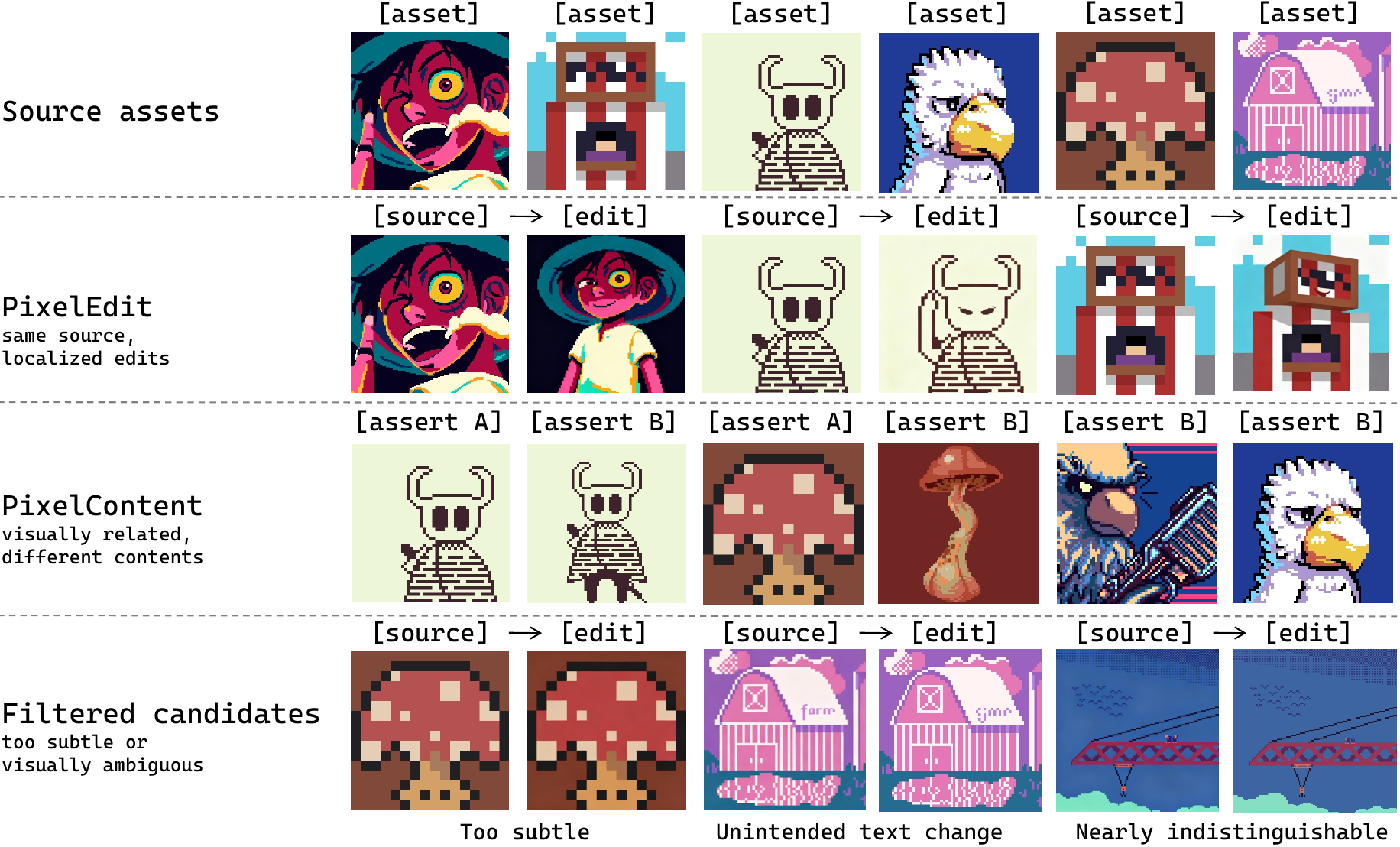}
\caption{
Representative pixel-art sources and constructed pairs.
The first row shows examples from the collected source pool.
The second row presents retained \textsc{PixelEdit} pairs, where the image on
the right is a localized edit of the source image on the left.
The third row presents \textsc{PixelContent} pairs consisting of visually
related but distinct source assets.
The final row shows candidate \textsc{PixelEdit} pairs removed during manual
review because their changes were too subtle, nearly indistinguishable, or
otherwise visually ambiguous.
}
\label{fig:app_pixel}
\end{figure*}

Figure~\ref{fig:app_pixel} summarizes the source pool, retained pairs, and
manual rejection process.
The rejected examples illustrate several cases removed from
\textsc{PixelEdit}, including changes that are too subtle, nearly
indistinguishable, or accompanied by an unintended modification.

\subsection{Scene and Region Sampling}
\label{app:scene_region_sampling}

After visual-pair curation, all five subsets share the same scene-generation
procedure.
A source pair provides two visual elements, one used as the majority reference
and the other used at the exception locations.
For each scene, we sample the grid dimensions, cell size, number of exception
cells, and their coordinates.
The majority element is rendered in every remaining cell, ensuring that the
scene contains a clear dominant reference.

When multiple exception cells are present, their coordinates are sampled with
a spread constraint.
This discourages unnecessary concentration in a single row or column and
reduces simple positional or grouping shortcuts.
The global exception locations are determined independently of the region
conditions subsequently used to instantiate the tasks.
Thus, a region is applied to an already defined scene-level exception set,
rather than determining where the exceptions are placed.

From the same scene, ROSE derives global, numeric-region, visual-region, and
exclusion-conditioned tasks.
Numeric regions may be row ranges, column ranges, or rectangles.
Visual regions use rectangular regions indicated directly in the image.
For exclusion tasks, the permitted region is the complement of a specified
row, column, or rectangle.

Local regions are sampled to include three diagnostically distinct relations
to the global exception set.
A \emph{Zero} region contains no exception cell, a \emph{Partial} region
contains a non-empty proper subset of the global exceptions, and an \emph{All}
region contains the complete global exception set.
These cases respectively test whether a model can abstain from action, filter
a global interpretation to a context-specific subset, and execute the complete
target set under a local instruction.

Exclusion regions are sampled with the same objective.
Depending on the scene and excluded area, the exclusion may remove no
exception, some exceptions, or all otherwise actionable exceptions.
Consequently, reusing the global count or global coordinate set is not a
reliable strategy across task contexts.

\subsection{Rendering and Cue-Visibility Controls}
\label{app:rendering_controls}

All visual elements are normalized before being placed into the grid.
Within a scene, the majority and exception elements use the same cell geometry,
padding policy, and rendering configuration.
Grid lines are drawn after the cell assets are placed, keeping row and column
boundaries visible and preserving the coordinate structure required by the
formal action protocol.

For glyph scenes, both characters use the same verified font and font size.
For emoji scenes, the selected provider assets are normalized to a common cell
extent while preserving their native style.
For pixel-art scenes, source assets are resized into fixed-size cells with
controlled padding so that differences in original image dimensions do not
directly reveal the exception.

Visual-region tasks use either a mask or an outline aligned with grid-cell
boundaries.
The cue specifies only the permitted region and does not mark the target cells.
For visually complex pixel-art scenes, a fixed boundary color may become
difficult to distinguish from the image content.
We therefore select a high-contrast boundary against the local image colors
and add a grid-aligned halo around the region boundary.
This keeps the cue visible without obscuring the visual elements or shifting
the intended cell coordinates.

Together, these rendering controls ensure that the visual source determines
the perceptual difficulty, while region cues remain interpretable and the
majority--exception distinction is not confounded by inconsistent scale,
padding, typeface, or boundary visibility.

\paragraph{Quality-control objective.}
The construction process is intentionally conservative.
Unsupported glyphs, inconsistent emoji sequences, invisible edits, ambiguous
pixel pairs, and unreliable visual cues are removed before benchmark
instantiation.
These filters reduce raw source volume, but make the resulting failures more
interpretable: the benchmark is designed to expose limitations in visual
reference inference, context-dependent selection, and symbolic execution
rather than artifacts of broken assets or ambiguous rendering.

\section{Prompts and Inference Details}
\label{app:prompt_inference}

\subsection{Shared Protocol Prompt}
\label{app:shared_prompt}

All evaluated models receive the same shared protocol instruction. The instruction specifies the grid indexing convention, restricts the response to the formal output space, and provides format examples only. It does not contain a solved task example.

\begin{quote}
\small\ttfamily
You are given a grid image. Rows are counted from top to bottom, and columns from left to right, both starting from 1. Output only the formal answer in the required format. Do not output any explanation.

\medskip
Allowed answer formats:

1) COUNT(n)

2) CLICK(Rr,Cc); CLICK(Rr,Cc); ...; DONE

3) CLICK(Rr,Cc); CLICK(Rr,Cc); ...; SUBMIT(n)

\medskip
Examples:

COUNT(3)

CLICK(R2,C5); CLICK(R4,C7); DONE

CLICK(R2,C5); CLICK(R4,C7); SUBMIT(2)
\end{quote}

\subsection{Task-Specific Prompt Templates}
\label{app:task_prompts}

The shared protocol instruction is followed by the image and a task-specific instruction. Numeric regions are instantiated using one of the following descriptions: \emph{in row $r$}, \emph{in rows $r_1$ to $r_2$}, \emph{in column $c$}, \emph{in columns $c_1$ to $c_2$}, or \emph{from row $r_1$ column $c_1$ to row $r_2$ column $c_2$}. The five task templates use the following prompts.

\paragraph{T1: Global counting.}
\begin{quote}
\small\ttfamily
Count the number of cells that are different from the majority in the whole grid. Answer only in the format: COUNT(n).
\end{quote}

\paragraph{T2: Local counting.}
\begin{quote}
\small\ttfamily
Count the number of cells that are different from the majority [REGION]. Answer only in the format: COUNT(n).
\end{quote}
Here, \texttt{[REGION]} is replaced by one of the numeric region descriptions defined above.

\paragraph{T3: Local clicking.}
\begin{quote}
\small\ttfamily
Click all cells that are different from the majority [REGION]. Answer only using CLICK(Rr,Cc); ...; DONE.
\end{quote}
The numeric region description is identical to that used for T2.

\paragraph{T4: Visual-region clicking.}
\begin{quote}
\small\ttfamily
Click all cells that are different from the majority inside the highlighted region. Answer only using CLICK(Rr,Cc); ...; DONE.
\end{quote}
The highlighted region is indicated directly in the image using either an outline box or a mask.

\paragraph{T5: Exclusion clicking with count submission.}
Depending on the excluded region type, the prompt takes one of the following forms.

\begin{quote}
\small\ttfamily
Click all cells that are different from the majority in all cells except row $r$, then submit the total number of clicked cells. Answer only using CLICK(Rr,Cc); ...; SUBMIT(n).
\end{quote}

\begin{quote}
\small\ttfamily
Click all cells that are different from the majority in all cells except column $c$, then submit the total number of clicked cells. Answer only using CLICK(Rr,Cc); ...; SUBMIT(n).
\end{quote}

\begin{quote}
\small\ttfamily
Click all cells that are different from the majority in all cells except the rectangle from row $r_1$ column $c_1$ to row $r_2$ column $c_2$, then submit the total number of clicked cells. Answer only using CLICK(Rr,Cc); ...; SUBMIT(n).
\end{quote}

\subsection{Model and Inference Configurations}
\label{app:inference_config}

We evaluate the nine MLLMs listed in Table~\ref{tab:main_results} using the model versions indicated by their reported names. Each benchmark item is evaluated independently in a single-turn request with no conversational history. The request contains the shared protocol instruction, one grid image, and the corresponding task-specific instruction, in that order.

Temperature is set to 0 for all models, and explicit reasoning or thinking modes are disabled where supported. Models are instructed to output only the formal answer, without an explanation. One successfully returned completion is retained and scored for each item. Request-level retries are used only when the API fails to return a response; successfully completed items are not resampled. The released code and run-configuration files provide the complete API identifiers and implementation-level settings.

\section{Formal Evaluation Definitions}
\label{app:formal_evaluation}

\subsection{Strict Parsing and Task Success}
\label{app:strict_parsing}

For an evaluation item $i$, let $\mathcal{G}_i$ denote the ground-truth coordinate set under the specified task condition, and let $n_i = |\mathcal{G}_i|$ denote the corresponding target count.
Depending on the task mode, the expected output takes one of the following forms:
\begin{equation*}
\texttt{COUNT}(n),
\end{equation*}
\begin{equation*}
\texttt{CLICK}(Rr,Cc);\,\ldots;\,\texttt{DONE},
\end{equation*}
or
\begin{equation*}
\texttt{CLICK}(Rr,Cc);\,\ldots;\,\texttt{SUBMIT}(n).
\end{equation*}

The parser is case-insensitive and permits whitespace inside formal tokens, but does not accept additional natural-language text.
Let $v_i \in \{0,1\}$ indicate whether the response satisfies the task-specific output grammar.

For click-applicable tasks, let $\mathcal{P}_i$ denote the set of unique parsed coordinates that fall within the valid grid.
We define a clean click response indicator as
\begin{equation*}
c_i
=
v_i
\cdot
\mathbf{1}\!\left[\text{no out-of-grid coordinate}\right]
\cdot
\mathbf{1}\!\left[\text{no repeated coordinate}\right].
\end{equation*}

For a counting task, PASS is defined as
\begin{equation*}
\operatorname{PASS}_i
=
v_i
\cdot
\mathbf{1}\!\left[\hat{n}_i=n_i\right],
\end{equation*}
where $\hat{n}_i$ is the predicted count.

For a click-only task, PASS requires an exact coordinate-set match:
\begin{equation*}
\operatorname{PASS}_i
=
c_i
\cdot
\mathbf{1}\!\left[\mathcal{P}_i=\mathcal{G}_i\right].
\end{equation*}
The order of click actions is ignored, but the predicted set must contain all and only the ground-truth coordinates.

For a click-and-submit task, PASS additionally requires the submitted count to match both the ground-truth count and the number of unique predicted clicks:
\begin{equation*}
\operatorname{PASS}_i
=
c_i
\cdot
\mathbf{1}\!\left[\mathcal{P}_i=\mathcal{G}_i\right]
\cdot
\mathbf{1}\!\left[\hat{n}_i=n_i\right]
\cdot
\mathbf{1}\!\left[\hat{n}_i=|\mathcal{P}_i|\right].
\end{equation*}

For zero-target items, the exact valid outputs are
\begin{equation*}
\texttt{COUNT}(0),\qquad
\texttt{DONE},\qquad
\texttt{SUBMIT}(0),
\end{equation*}
for counting, click-only, and click-and-submit tasks, respectively.

VALID measures only whether the response satisfies the required formal grammar:
\begin{equation*}
\operatorname{VALID}_i=v_i.
\end{equation*}
It does not require the predicted count or coordinate set to be correct.
A response may therefore be grammar-valid while containing an out-of-grid coordinate, a repeated coordinate, a region violation, or an incorrect target set.

\subsection{Partial-Credit Metrics}
\label{app:partial_credit}

For counting tasks, SOFT is based on the absolute count error:
\begin{equation*}
S_i^{\mathrm{cnt}}
=
\frac{1}{1+|\hat{n}_i-n_i|}.
\end{equation*}
If no count can be parsed, the score is set to zero.

For click-applicable tasks, coordinate-set F1 is computed over the unique in-grid predicted coordinates:
\begin{equation*}
F_i
=
\frac{
2|\mathcal{P}_i\cap\mathcal{G}_i|
}{
|\mathcal{P}_i|+|\mathcal{G}_i|
}.
\end{equation*}
When both $\mathcal{P}_i$ and $\mathcal{G}_i$ are empty, $F_i$ is defined as 1.
When exactly one of the two sets is empty, $F_i$ is defined as 0.

We use a strict coordinate-set F1 that also accounts for formal action validity:
\begin{equation*}
F_i^{\mathrm{strict}}
=
\begin{cases}
F_i, & c_i=1,\\
0, & c_i=0.
\end{cases}
\end{equation*}
Thus, a malformed output, an out-of-grid coordinate, or a repeated coordinate sets the strict coordinate F1 to zero.

For click-only tasks, SOFT is defined as
\begin{equation*}
S_i^{\mathrm{clk}}
=
F_i^{\mathrm{strict}}.
\end{equation*}

For click-and-submit tasks, let
\begin{equation*}
C_i
=
\mathbf{1}\!\left[\hat{n}_i=|\mathcal{P}_i|\right]
\end{equation*}
denote consistency between the submitted count and the number of unique in-grid predicted clicks.
If the submitted count cannot be parsed, $C_i$ is set to zero.
The corresponding SOFT score is
\begin{equation*}
S_i^{\mathrm{cs}}
=
\frac{
F_i^{\mathrm{strict}}
+
S_i^{\mathrm{cnt}}
+
C_i
}{3}.
\end{equation*}

The unified item-level SOFT score is selected according to the task mode:
\begin{equation*}
S_i
=
\begin{cases}
S_i^{\mathrm{cnt}}, & \text{for counting tasks},\\
S_i^{\mathrm{clk}}, & \text{for click-only tasks},\\
S_i^{\mathrm{cs}}, & \text{for click-and-submit tasks}.
\end{cases}
\end{equation*}

C-F1 is the mean strict coordinate-set F1 over all click-applicable items:
\begin{equation*}
\operatorname{C\mbox{-}F1}
=
\frac{1}{|\mathcal{D}_{\mathrm{clk}}|}
\sum_{i\in\mathcal{D}_{\mathrm{clk}}}
F_i^{\mathrm{strict}},
\end{equation*}
where $\mathcal{D}_{\mathrm{clk}}$ contains both click-only and click-and-submit items.

\subsection{Region Compliance and Aggregation}
\label{app:region_aggregation}

Let $\mathcal{R}_i$ denote the set of grid cells permitted by the task condition.
For a click-applicable item, the region-violation rate is defined as
\begin{equation*}
E_i^{\mathrm{reg}}
=
\begin{cases}
\displaystyle
\frac{
|\mathcal{P}_i\setminus\mathcal{R}_i|
}{
|\mathcal{P}_i|
},
& |\mathcal{P}_i|>0,\\[10pt]
0,
& |\mathcal{P}_i|=0.
\end{cases}
\end{equation*}

R-OK is the complement of the mean region-violation rate:
\begin{equation*}
\operatorname{R\mbox{-}OK}
=
1
-
\frac{1}{|\mathcal{D}_{\mathrm{clk}}|}
\sum_{i\in\mathcal{D}_{\mathrm{clk}}}
E_i^{\mathrm{reg}}.
\end{equation*}
R-OK measures whether predicted in-grid clicks remain within the required region.
It does not measure whether the correct target coordinates were selected.
An empty predicted set therefore incurs no region violation, while its localization accuracy is captured separately by PASS and C-F1.

For a zero-target click item, an empty predicted set receives coordinate-set F1 equal to 1.
A non-empty predicted set receives coordinate-set F1 equal to 0 because the ground-truth set is empty.
Strict PASS additionally requires the correct terminal output, namely \texttt{DONE} or \texttt{SUBMIT}(0), depending on the task mode.

Unless otherwise stated, each metric is first averaged over the applicable items within each visual subset.
The final model-level result is then computed as an equal macro average over the five subsets.
Let $M_s$ denote the value of metric $M$ on visual subset $s$.
The reported macro average is
\begin{equation*}
M_{\mathrm{macro}}
=
\frac{1}{5}
\sum_{s=1}^{5}
M_s.
\end{equation*}

For task-specific results, $M_s$ is computed only over items belonging to the corresponding task template.
For click diagnostics such as C-F1 and R-OK, $M_s$ is computed only over click-applicable items.

\section{Controlled Bridge Tasks}
\label{app:controlled_bridges}

The main ROSE benchmark compares several tasks derived from the same visual
scene, but some aggregate differences may still combine multiple sources of
difficulty. In particular, a drop from counting to region-conditioned clicking
may reflect both coordinate grounding and context-dependent target selection.
Similarly, the original local-count and local-click templates use independently
sampled numeric regions, so their aggregate difference is not a strictly
instance-matched comparison.

We therefore construct two additional bridge tasks from the official ROSE test
split. Both bridges reuse existing scenes and image assets without introducing
new visual examples. The first inserts global coordinate localization between
global counting and region-conditioned action. The second creates an exactly
matched local count--click pair in which the image, numeric region, and regional
target set are held fixed. These controls are used only for diagnostic analysis
and do not change the main ROSE benchmark or its official five task templates.

All bridge items are evaluated using the same shared protocol instruction,
single-turn inference procedure, model configurations, and strict parser as the
main benchmark. Each item is queried independently, with no conversational
history or transfer of intermediate model states between paired tasks.
Consequently, the conditioned results below measure behavioral consistency
across paired task contexts rather than literal hidden-state transfer between
requests.

\begin{table}[t]
\centering
\small
\setlength{\tabcolsep}{5.5pt}
\begin{tabular}{lrr}
\toprule
\textbf{Subset} &
\textbf{Global-Click} &
\textbf{Matched Local Count} \\
\midrule
\textsc{ChineseGlyph} & 301 & 301 \\
\textsc{EmojiStyle}   & 221 & 221 \\
\textsc{EmojiContent} & 221 & 221 \\
\textsc{PixelEdit}    & 221 & 221 \\
\textsc{PixelContent} & 148 & 148 \\
\midrule
\textbf{Total}        & 1,112 & 1,112 \\
\bottomrule
\end{tabular}
\caption{
Statistics of the two controlled bridge tasks on the official test split.
Each Global-Click item is derived from one original global-count item, while
each Matched Local Count item is derived from one original numeric local-click
item.
}
\label{tab:bridge_statistics}
\end{table}

\subsection{Global Count-to-Click Bridge}
\label{app:global_click_bridge}

\paragraph{Motivation.}
A correct global count does not necessarily imply that a model has localized
every exception cell correctly. A model may recover the correct cardinality
while confusing the underlying coordinates, for example by missing one target
and introducing one false positive. Therefore, the difference between global
counting and region-conditioned clicking may contain two distinct components:
converting a visually inferred target set into exact coordinates, and filtering
that target set under a new region context.

To separate these components, we introduce an intermediate global-click task,
denoted \textbf{G-Clk}. It uses the same uncued image and the same global
exception set as the original global-count task, denoted \textbf{G-Cnt}, but
requires the model to return the complete set of exception coordinates rather
than only their number.

\paragraph{Construction.}
For every original \texttt{T1\_COUNT\_GLOBAL} item in the official test split,
we derive one \texttt{T1B\_CLICK\_GLOBAL} item. The derived item preserves the
following fields from its source item:

\begin{itemize}
    \item the scene and rendered image;
    \item the majority and exception visual elements;
    \item the full grid dimensions;
    \item the complete global exception set;
    \item the official scene-level split assignment; and
    \item the shared protocol instruction and row--column indexing convention.
\end{itemize}

The region remains the full grid, with no numeric or visual region cue:
\texttt{region\_type=GLOBAL} and \texttt{cue\_type=NONE}. The only task-level
change is that the required response operation is changed from global counting
to global coordinate selection. The task-specific instruction is

\begin{quote}
\small\ttfamily
Click all cells that are different from the majority in the whole grid.
Answer only using CLICK(Rr,Cc); ...; DONE.
\end{quote}

If the global exception set is
\[
O = \{(r,c)\in G : v_{r,c} \neq v^\star\},
\]
then the ground-truth click set for G-Clk is exactly \(O\). The bridge item
retains the identifier of its source G-Cnt item, allowing the two independently
evaluated predictions to be paired exactly at the item or scene level.

\paragraph{Evaluation.}
G-Clk uses the same strict click evaluator as the main ROSE action tasks.
Click order is ignored, but strict PASS requires that the predicted coordinate
set contain every ground-truth cell exactly once and no additional cells.
Malformed outputs, out-of-grid coordinates, and repeated coordinates are
treated as failures.

We additionally report the following bridge diagnostics:

\begin{itemize}
    \item \textbf{G-Clk}: strict global-click PASS over all derived items;
    \item \textbf{G-Clk\(^{\dagger}\)}: global-click PASS restricted to paired
    scenes where the same model's independent G-Cnt prediction is correct;
    \item \textbf{Card}: strict clicked-cardinality accuracy, requiring a
    grammar-valid response with no invalid or repeated coordinates and with the
    number of unique predicted clicks equal to the ground-truth cardinality;
    \item \textbf{Loc.\(\mid\)Card}: exact coordinate-set accuracy restricted
    to items with correct strict clicked cardinality; and
    \item \textbf{C-F1}: strict coordinate-set F1, using the same definition as
    in the main evaluation.
\end{itemize}

Let \(g_i\in\{0,1\}\) denote G-Cnt PASS and \(b_i\in\{0,1\}\) denote G-Clk
PASS for a paired scene \(i\). The conditioned bridge score is

\[
\mathrm{G\mbox{-}Clk}^{\dagger}
=
\frac{\sum_i g_i b_i}{\sum_i g_i}.
\]

This conditioned metric controls for recovery of the correct global
cardinality, but it does not assume that a correct count proves exact
localization. The unconditional G-Clk result directly measures this missing
localization step.

\paragraph{Interpretation.}
The resulting three-stage comparison
\[
\mathrm{G\mbox{-}Cnt}
\rightarrow
\mathrm{G\mbox{-}Clk}
\rightarrow
\mathrm{V\mbox{-}Clk}
\]
decomposes the original counting-to-action gap. The first transition introduces
exact coordinate localization while keeping the full-grid context unchanged.
The second transition introduces a region-conditioned selection rule on top of
coordinate-level action. A decrease from G-Cnt to G-Clk therefore measures the
count-to-coordinate component, whereas an additional decrease from G-Clk to
visual-region clicking indicates difficulty in rebinding the globally inferred
target set to the current region context.

\subsection{Matched Local Count-to-Click Bridge}
\label{app:matched_local_bridge}

\paragraph{Motivation.}
The original local-count task and numeric local-click task are both derived from
the same scene collection, but their regions are sampled independently.
Consequently, their aggregate score difference may partly reflect variation in
region difficulty, target composition, or the number of regional exceptions.
A stricter comparison requires holding the image, region, and regional target
set fixed while changing only the required output operation.

We therefore construct a matched local-count task, denoted \textbf{mL-Cnt},
from every original numeric local-click item, denoted \textbf{L-Clk}. Each pair
asks about exactly the same regional exception set, but mL-Cnt requires its
cardinality whereas L-Clk requires its coordinates.

\paragraph{Construction.}
For every original \texttt{T3\_CLICK\_LOCAL\_NUMERIC} item in the official test
split, we derive one
\texttt{T2B\_COUNT\_LOCAL\_MATCHED} item. The derived count item preserves:

\begin{itemize}
    \item the same scene and image;
    \item the same numeric region type;
    \item the same row, column, or rectangle parameters;
    \item the same global exception set;
    \item the same regional exception set;
    \item the same Zero, Partial, or All region case; and
    \item the same split assignment.
\end{itemize}

The supported numeric region types are row ranges, column ranges, and
rectangles. If the original L-Clk item defines a permitted region \(R_i\), its
target set is

\[
T_i = O_i \cap R_i,
\]

where \(O_i\) is the global exception set of the corresponding scene. The
original L-Clk ground truth is the coordinate set \(T_i\), while the derived
mL-Cnt ground truth is its cardinality \(|T_i|\).

The derived task uses the same numeric region wording as the source local-click
item. Its task-specific instruction has the form

\begin{quote}
\small\ttfamily
Count the number of cells that are different from the majority [REGION].
Answer only in the format: COUNT(n).
\end{quote}

Here, \texttt{[REGION]} is instantiated from the preserved region type and
parameters, for example ``in row \(r\),'' ``in columns \(c_1\) to \(c_2\),''
or ``from row \(r_1\) column \(c_1\) to row \(r_2\) column \(c_2\).'' Each
derived item stores the identifier of its source L-Clk item, enabling exact
one-to-one pairing during evaluation.

The construction script verifies that the source item is a numeric local-click
task, that it contains no visual region cue, and that its recorded target count
equals the size of its regional target set. It also verifies that no source
local-click item is used more than once and that the derived item preserves the
same scene, image, region parameters, and target cardinality.

\paragraph{Region cases.}
The matched pairs are grouped according to the relation between the regional
target set \(T_i\) and the global exception set \(O_i\):

\begin{itemize}
    \item \textbf{Zero}: \(T_i=\varnothing\), so the correct count is
    \texttt{COUNT(0)} and the paired click action is \texttt{DONE};
    \item \textbf{Partial}: \(\varnothing \subset T_i \subset O_i\), so the
    region contains a non-empty proper subset of the global exceptions; and
    \item \textbf{All}: \(T_i=O_i\), so every global exception lies inside the
    numeric region.
\end{itemize}

This breakdown distinguishes failures of abstention from failures of
region-specific filtering and full-set localization.

\paragraph{Evaluation.}
The new mL-Cnt predictions are scored with the standard ROSE count parser and
strict COUNT PASS definition. The paired L-Clk results are not rerun or
re-evaluated under a different protocol; they are loaded from the original
per-item ROSE evaluation and joined using the stored source local-click item
identifier. The evaluator verifies that the mL-Cnt target count and the
cardinality of the paired L-Clk target set agree.

We report:

\begin{itemize}
    \item \textbf{mL-Cnt}: strict PASS on the derived matched count task;
    \item \textbf{L-Clk}: strict PASS on the corresponding original numeric
    local-click items;
    \item \textbf{L-Clk\(^{\dagger}\)}: L-Clk PASS restricted to pairs where
    the independently queried mL-Cnt response is correct; and
    \item \textbf{Fail\(^{\dagger}\)}:
    \(1-\mathrm{L\mbox{-}Clk}^{\dagger}\), the fraction of correctly counted
    matched regions that are not converted into an exact click set.
\end{itemize}

Let \(m_i\in\{0,1\}\) denote mL-Cnt PASS and
\(\ell_i\in\{0,1\}\) denote paired L-Clk PASS. The conditioned action score is

\[
\mathrm{L\mbox{-}Clk}^{\dagger}
=
\frac{\sum_i m_i \ell_i}{\sum_i m_i},
\]

and the corresponding conditional failure rate is

\[
\mathrm{Fail}^{\dagger}
=
1-\mathrm{L\mbox{-}Clk}^{\dagger}.
\]

Because the two tasks are queried independently, this quantity should be
interpreted as matched cross-task consistency: among regions for which the
model returns the correct target cardinality under the count prompt, how often
does it also return the exact target coordinate set under the paired click
prompt?

\paragraph{Interpretation.}
The matched design removes variation in image content, region geometry, region
size, and target composition. A failure under
L-Clk\(^{\dagger}\) therefore cannot be attributed to the paired count and click
tasks referring to different regions. It instead indicates that correct
regional cardinality is not sufficient for exact coordinate-level execution.
The Zero case tests whether a model can suppress all actions after determining
that no valid target exists. The Partial case tests whether it can select a
region-specific subset rather than reverting to the full-scene exception set.
The All case tests exact localization when regional filtering does not remove
any global target.

\subsection{Aggregation and Reproducibility}
\label{app:bridge_aggregation}

All bridge metrics are first computed separately within each of the five visual
subsets. The reported model-level result is then the equal macro average

\[
M_{\mathrm{macro}}
=
\frac{1}{5}\sum_{s=1}^{5} M_s.
\]

The same aggregation rule is applied to unconditional, conditioned, and
region-case-specific results. Thus, larger subsets do not dominate the reported
bridge scores.

The bridge datasets are generated deterministically from the official ROSE
split files. Each derived item stores its source item identifier and preserves
the relevant scene, image, region, and target metadata. API inference uses the
same generic ROSE runner as the main benchmark, and both bridge evaluators
reuse the standard per-item parser and task-success implementation. The
released generation scripts, manifests, per-item paired evaluations, and
model-level summaries provide the full data lineage from each bridge item back
to its original ROSE task.

\section{Additional Diagnostics}
\label{app:additional_diagnostics}

\subsection{Click Error Taxonomy}
\label{app:click_error_taxonomy}

To characterize failures on click-applicable tasks, we assign each failed prediction to one mutually exclusive category using the following priority order.

\paragraph{Protocol or invalid output.}
This category includes responses with invalid output grammar, out-of-grid coordinates, repeated coordinates, or an inconsistent \texttt{SUBMIT} count.

\paragraph{Region violation.}
If the response is otherwise valid, it is assigned to this category when at least one predicted in-grid coordinate lies outside the region specified by the task.

\paragraph{Cardinality error.}
If the response is valid and region-compliant, it is assigned to this category when the number of unique predicted coordinates differs from the ground-truth target count.

\paragraph{Wrong location.}
The remaining failed responses have valid and region-compliant outputs with the correct click cardinality, but the predicted coordinate set does not exactly match the ground-truth set.

For each model, category proportions are computed over failed click-applicable items only. Because the categories are assigned sequentially, every failed prediction contributes to exactly one category and the reported proportions sum to 100\%.

\section{Additional Experiments}

\begin{figure*}[t]
\centering
\includegraphics[width=0.45\textwidth]{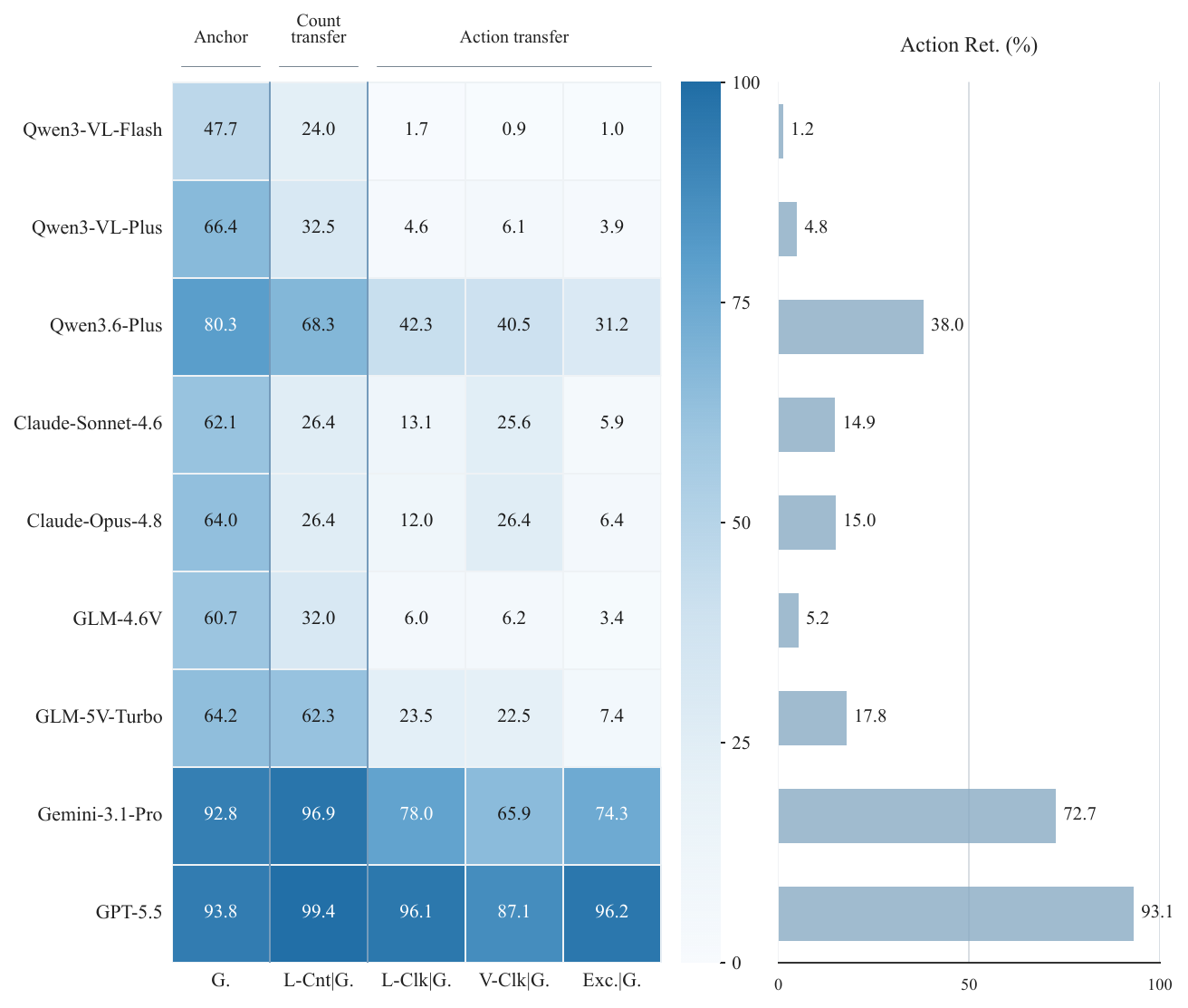}
\caption{
Scene-coupled consistency under correct global counting.
G. denotes global-count PASS.
L-Cnt$|$G., L-Clk$|$G., V-Clk$|$G., and Exc.$|$G. report the corresponding task accuracies evaluated only on scenes where the same model first solves G. correctly.
Action Ret. is the mean of L-Clk$|$G., V-Clk$|$G., and Exc.$|$G.
}
\label{fig:scene_consistency}
\end{figure*}

\paragraph{Does a correct anchor scene interpretation transfer across contexts?}
Figure~\ref{fig:scene_consistency} sharpens the perception-to-action gap by conditioning on scenes where the same model already solves G.\ correctly.
Under this control, the remaining errors can no longer be explained simply by failure to identify the odd cells in the original scene.
Yet a substantial gap still remains: several models retain much higher L-Cnt$|$G.\ than conditioned action accuracy, indicating that the bottleneck lies less in perception itself than in rebinding a correct scene interpretation to a new task context and converting it into an exact symbolic action.
This is particularly clear for Qwen3.6-Plus and GLM-5V-Turbo, while the Claude models suggest that explicit visual-region cues help more than pure local clicking or exclusion.
Even for Gemini-3.1-Pro and GPT-5.5, V-Clk$|$G.\ remains the weakest conditioned action column.
Overall, ROSE isolates a stricter bottleneck than global odd-cell detection: whether a correct scene-level interpretation can survive context change and be carried through to coordinate-level action.

\begin{figure*}[t]
\centering
\includegraphics[width=\textwidth]{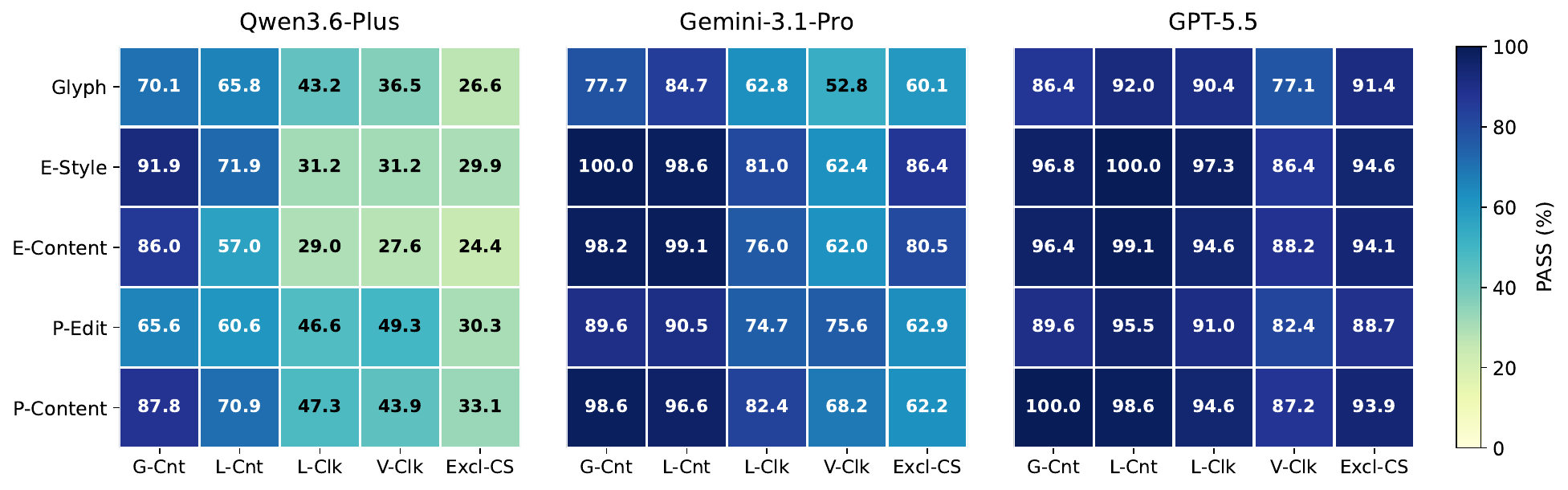}
\caption{
Subset--template PASS heatmaps for representative models.
Rows denote visual subsets and columns denote task templates.
All panels share the same color scale.
}
\label{fig:subset_template_heatmap}
\end{figure*}

\paragraph{Is the action gap tied to specific visual sources?}
Figure~\ref{fig:subset_template_heatmap} examines whether the perception-to-action gap is driven by a particular visual subset or appears consistently across different sources of visual variation.
Across representative models, counting templates generally remain stronger than action templates within the same visual subset.
For Qwen3.6-Plus, G-Cnt and L-Cnt are relatively high across most subsets, yet L-Clk, V-Clk, and especially Excl-CS drop substantially on all five visual sources.
This indicates that its action weakness is not caused by a single difficult subset, but by the additional requirement of converting the perceived odd cells into constrained coordinate-level actions.

The same pattern remains visible, though less severely, for stronger models.
Gemini-3.1-Pro nearly solves counting on Emoji and Pixel subsets, but V-Clk remains consistently lower than the corresponding counting scores.
GPT-5.5 achieves high performance across nearly all cells, confirming that the tasks are solvable, yet V-Clk is still its weakest template on multiple subsets.
Together, these heatmaps suggest that ROSE's difficulty is not merely a property of Chinese glyphs, emoji renderings, or pixel-level edits alone.
Instead, the main bottleneck emerges when the same visual evidence must be reinterpreted under a task-specific context and expressed as exact symbolic action.

\begin{figure*}[t]
\centering
\includegraphics[width=\textwidth]{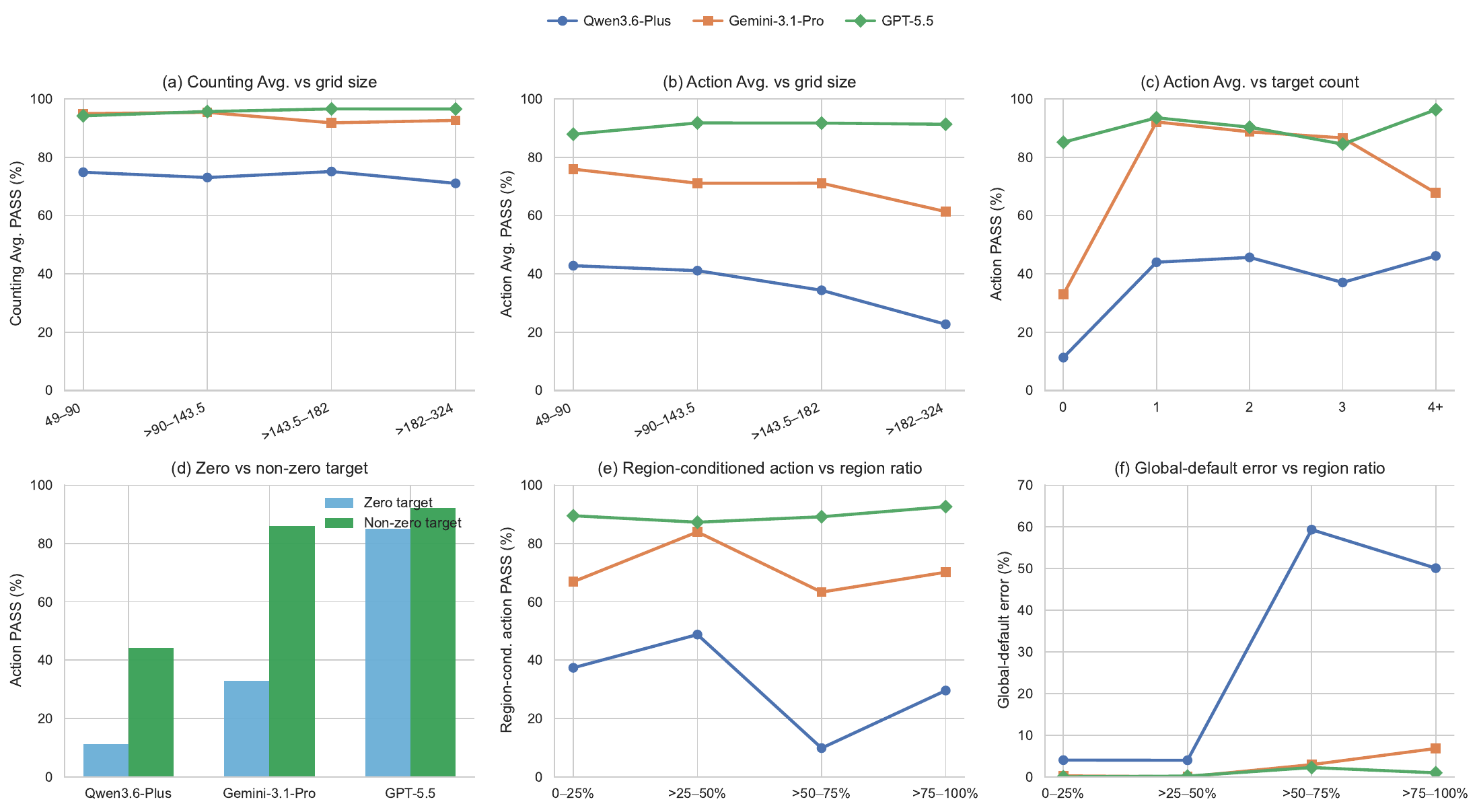}
\caption{
Difficulty scaling in ROSE.
(a,b) Counting and action performance across grid-size bins.
(c) Action performance across task target counts.
(d) Action performance on zero-target and non-zero-target cases.
(e) Region-conditioned action performance across region-area ratios.
(f) Global-default error across region-area ratios.
}
\label{fig:difficulty_scaling}
\end{figure*}

\paragraph{How does difficulty scale with scene and target complexity?}
Figure~\ref{fig:difficulty_scaling} examines whether ROSE difficulty can be explained by simple instance-level factors such as grid size, target count, or region size.
The results suggest that scene scale alone is not sufficient to explain the benchmark.
In Figure~\ref{fig:difficulty_scaling}(a), Counting Avg.\ remains relatively stable as grid size increases: Qwen3.6-Plus varies only from 74.8\% to 71.0\%, Gemini-3.1-Pro from 95.0\% to 92.6\%, and GPT-5.5 from 94.2\% to 96.5\%.
However, Figure~\ref{fig:difficulty_scaling}(b) shows that action performance is more sensitive to the same increase in grid size.
Qwen3.6-Plus drops from 42.8\% Action Avg.\ in the smallest grid bin to 22.7\% in the largest bin, while Gemini-3.1-Pro drops from 75.9\% to 61.3\%.
GPT-5.5 remains much more stable, staying above 87.9\% across all grid-size bins.
This contrast indicates that larger scenes do not merely make the odd cells impossible to perceive; they more strongly stress the conversion from a visual decision into exact coordinate-level action.

Target-count scaling reveals a different bottleneck.
Figure~\ref{fig:difficulty_scaling}(c) does not show a simple monotonic degradation as the number of required clicks increases.
Instead, the clearest discontinuity is between zero-target and non-zero-target cases.
As shown in Figure~\ref{fig:difficulty_scaling}(d), Qwen3.6-Plus reaches only 11.3\% Action PASS when the correct action contains no target, compared with 44.1\% on non-zero-target cases.
The gap is even more pronounced for Gemini-3.1-Pro, which rises from 32.9\% on zero-target cases to 86.0\% on non-zero-target cases.
GPT-5.5 also shows a smaller version of this effect, from 85.1\% to 92.3\%.
Thus, ROSE tests not only whether a model can select the right cells, but also whether it can abstain from clicking when the current region or exclusion condition leaves no valid target.

Finally, Figures~\ref{fig:difficulty_scaling}(e,f) show that region-conditioned difficulty is not determined by region area ratio alone.
GPT-5.5 remains robust across region-ratio bins, with region-conditioned action performance between 87.3\% and 92.6\% and near-zero global-default error.
Gemini-3.1-Pro is weaker but still keeps global-default error low, rising only to 6.8\% in the largest-ratio bin.
In contrast, Qwen3.6-Plus becomes highly unstable in larger-ratio regimes: its region-conditioned action score falls to 9.8\% in the 50--75\% bin, while its global-default error rises to 59.3\%; in the 75--100\% bin, global-default error remains 50.1\%.
These errors indicate a specific failure mode in which the model falls back to the full-scene odd set or count instead of applying the current region constraint.
Overall, ROSE difficulty is structured rather than merely size-driven: counting remains comparatively robust, whereas action tasks expose sensitivity to grid scale, abstention, and context-dependent filtering.

\end{document}